\pdfoutput=1

\documentclass[11pt]{article}

\usepackage{acl}

\usepackage{times}
\usepackage{latexsym}

\usepackage{hyperref}
\usepackage{url}
\usepackage{tabularx}
\usepackage{makecell}
\usepackage{multirow}
\usepackage{booktabs}
\usepackage{lipsum}
\usepackage{adjustbox}
\usepackage{ragged2e} 
\usepackage{bbding}
\usepackage{enumitem}
\usepackage{dsfont}
\usepackage{amsmath}
\usepackage{amssymb}
\usepackage{wrapfig}
\usepackage{color}
\usepackage{soul}
\usepackage{lineno}
\usepackage{cleveref}
\usepackage{tabularray}
\usepackage{array}
\crefname{section}{§}{§§}
\Crefname{section}{§}{§§}
\usepackage{subcaption}
\usepackage{graphicx}

\usepackage[T1]{fontenc}

\usepackage[utf8]{inputenc}

\usepackage{microtype}

\usepackage{inconsolata}

\title{Instructing Large Language Models to \\ Identify and Ignore Irrelevant Conditions}

\author{Zhenyu Wu\textsuperscript{\rm 1,2}, Chao Shen\textsuperscript{\rm 1}\thanks{Corresponding author}, Meng Jiang\textsuperscript{\rm 2} \\
        \textsuperscript{\rm 1}Xi’an Jiaotong University, \textsuperscript{\rm 2}University of Notre Dame \\
        zhenyuwu@stu.xjtu.edu.cn, chaoshen@xjtu.edu.cn, mjiang2@nd.edu}

\begin{document}
\maketitle
\begin{abstract}
Math word problem (MWP) solving requires generating a reasoning path based on a given problem description that often contains \emph{irrelevant conditions}.
Existing chain-of-thought (CoT) prompting methods elicited multi-step reasoning abilities of large language models (LLMs) to solve MWPs.
However, they were seriously confused by the irrelevant conditions, resulting in low accuracy.
In this paper, we propose a novel approach named I$^3$C that instructs LLMs to identify and ignore irrelevant conditions.
It identifies a set of irrelevant condition candidates that have a weak semantic relevance with the question.
Then it prompts LLMs to verify the irrelevant conditions.
Lastly it instructs the LLMs with the verification on relevant and irrelevant conditions to avoid confusion and improve reasoning paths.
Moreover, we propose to select (problem, reasoning paths) pairs as demonstrations to enhance I$^3$C with few-shot reasoning. We develop I$^3$C-Select that selects the most confusing problems based on the semantic relevance measurement.
We conduct extensive experiments on eight MWP datasets.
I$^3$C can be combined with any CoT prompting methods to improve the performance of solving MWPs.
Notably, with GPT-3.5-Turbo and I$^3$C-Select, we achieve an accuracy of $96.0$ and $94.1$ on GSM-IC2-1K and GSM-ICM-1K, respectively, significantly outperforming the state-of-the-art few-shot prompting method Complex-CoT by $+11.7$ and $+11.1$.
Our implementation is made publicly available at \url{https://wzy6642.github.io/I3C.github.io/}.
\end{abstract}

\section{Introduction}
\label{sec:intro}
\begin{figure*}[t]
	\centering
    \begin{subfigure}{1.0\textwidth}
	\includegraphics[width=1.0\textwidth]{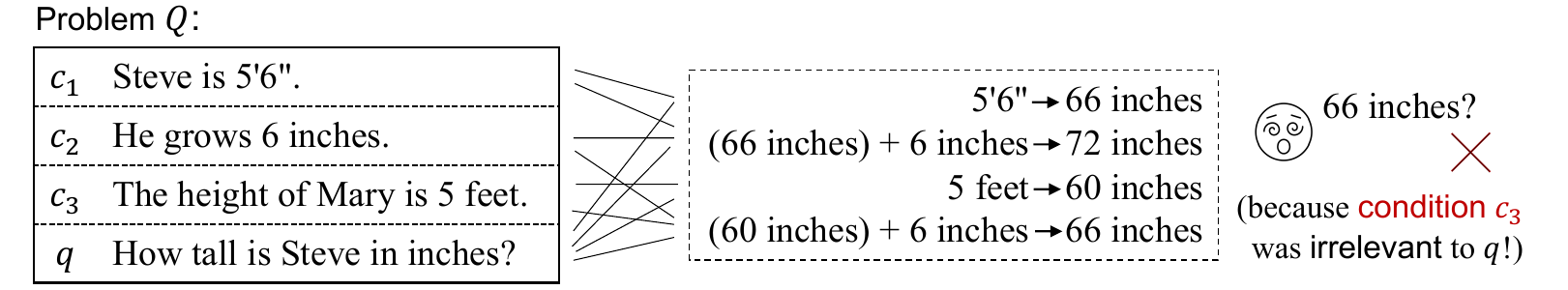}
    \caption{Existing CoT prompting methods were confused by irrelevant conditions in math word problems and gave wrong answers.}
    \label{fig:framework_1}
    \end{subfigure}
    \begin{subfigure}{1.0\textwidth}
	\includegraphics[width=1.0\textwidth]{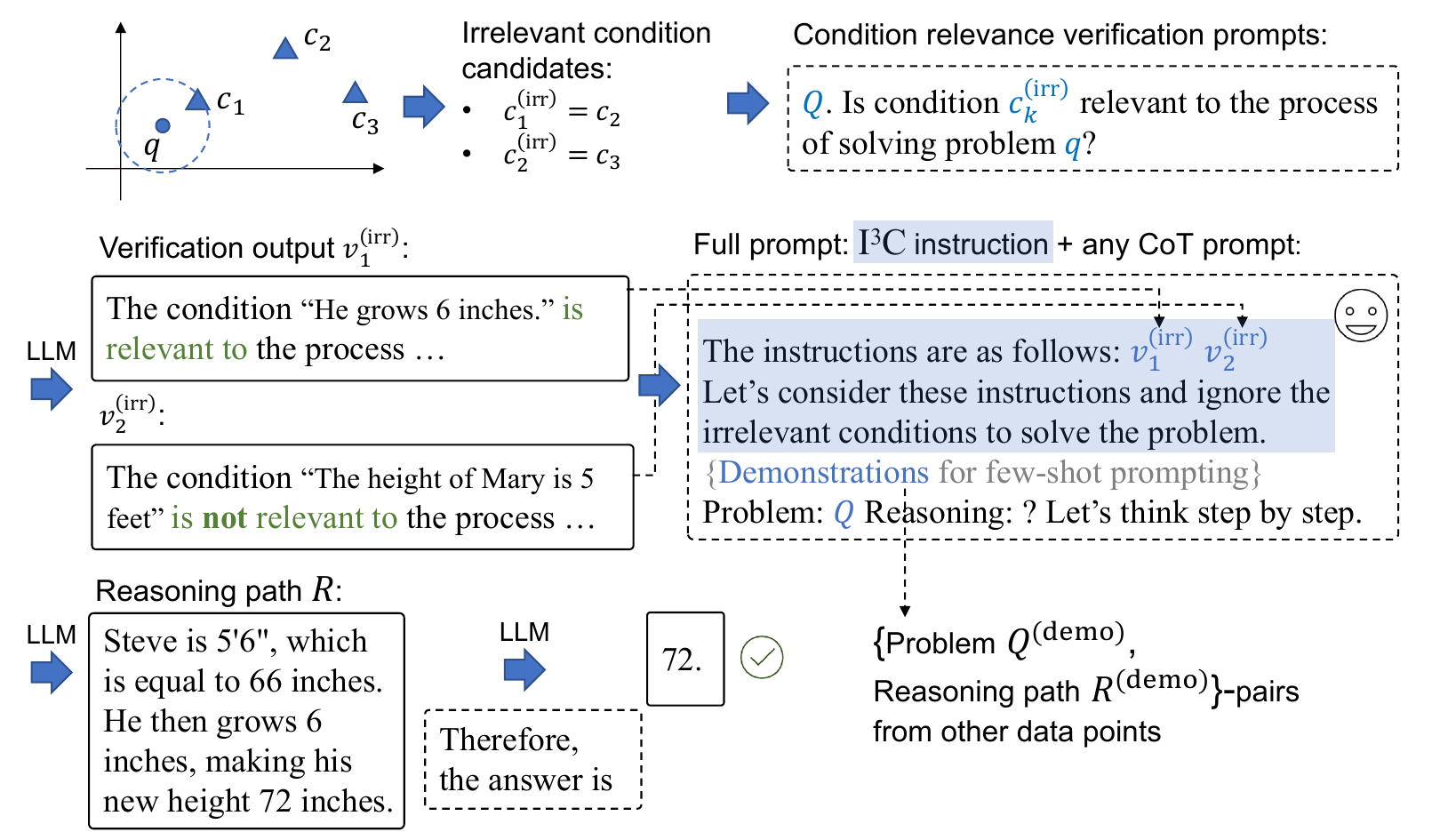}
	\caption{I$^3$C performs three steps: (1) Identify irrelevant condition candidates by encoding and condition-question similarity scoring; (2) Use LLMs to verify if the candidates are relevant; (3) Leverage the verifications (and demonstrations) to generate accurate reasoning paths and find correct answers.}
    \label{fig:framework_2}
    \end{subfigure}
    \caption{The proposed I$^3$C approach instructs LLMs to \underline{I}dentify and \underline{I}gnore \underline{I}rrelevant \underline{C}onditions.}
	\label{fig:framework}
\end{figure*}

Math word problem (MWP) solving is a task of developing algorithms to generate a reasoning path towards an unknown quantity based on a problem description.
This task is challenging as it requires mathematical understanding and multi-step reasoning abilities.
Chain-of-thought (CoT) prompting methods were able to guide large language models (LLMs) to perform complex multi-step reasoning \citep{Kojima-2022-CoT,wang-2023-ps}.
Adding demonstrations created manually \cite{wei-2022-cot} or retrieved from a large training set \cite{fu-2023-complexity} in CoT prompts enabled few-shot in-context learning and improved accuracy.
However, \citeauthor{freda-2023-gsm8k-ic} found that existing CoT prompting methods could be seriously confused by irrelevant conditions which are specifications or data presented in a problem that are unrelated to the solution \citep{kellogg-2016}. For example, as shown in Figure~\ref{fig:framework_1}, the third condition ``\emph{The height of Mary is 5 feet.}'' was irrelevant to the final question and misled the reasoning and prediction.
\citeauthor{freda-2023-gsm8k-ic} added a plain instruction ``\emph{Feel free to ignore irrelevant conditions in the problem description.}'' in the prompts, but the LLMs could not effectively ignore them in the problem solving process because they were not identified or specified in the instruction.

Improving the reasoning on MWPs that have irrelevant conditions is non-trivial. Self-consistency \citep{wang-2023-selfconsistency} was proposed to repeatedly solve a problem multiple times (e.g., 10 times) and employ a majority vote strategy to determine the most consistent answer as the final answer. However, it was computationally expensive and still confused by the irrelevant conditions. Moreover, the demonstrations would have to be re-designed to obtain the few-shot learning ability of identifying and ignoring the irrelevance, compared to those in \citep{wei-2022-cot,zhang-2023-autocot}.

In this paper, we propose a novel approach, I$^3$C, to instruct LLMs to explicitly \underline{I}dentify and \underline{I}gnore \underline{I}rrelevant \underline{C}onditions in the mathematical reasoning process. It creates effective instructions that can be added to any CoT prompts to improve their generated reasoning paths. 
Unlike self-consistency, I$^3$C does not prompt LLMs multiple times. 
Its advanced variant, I$^3$C-Select, uses the most confusing problems and their generated reasoning paths as demonstrations for few-shot learning.

First, we quantify the semantic relevance of each condition $c_i$ in a MWP $Q=[\{c_i\},q]$. Specifically, we use a language model like SimCSE~\citep{gao-2021-simcse} to encode the conditions $\{c_i\}$ and question sentence $q$. The semantic relevance is lower if the condition's encoding is more distant from the encodings of question and other conditions, as shown in Figure~\ref{fig:framework_2}. Then we identify a set of irrelevant condition candidates, like $c_2$ and $c_3$ in this example, and we denote them by $\{c^{\text{(irr)}}_k\}$.

Next we use an LLM to verify if the candidates are indeed irrelevant. For each candidate $c^{\text{(irr)}}_k$, the verification prompt is a natural language question consisted of itself, $Q$, and $q$. The verification output usually has the explicit answers ``\emph{... is (not) relevant to ...}'', denoted by $v^{\text{(irr)}}_k$.

Finally we put all the verification outputs $\{v^{\text{(irr)}}_k\}$ to create a novel instruction which helps the LLM to \underline{i}dentify and \underline{i}gnore \underline{i}rrelevant \underline{c}onditions in the problem description, so-called I$^3$C. The I$^3$C instruction is a plug-and-play module and can be added to any CoT prompting methods to help LLMs avoid confusion and improve generated reasoning paths.

To enable few-shot in-context learning, we further develop I$^3$C-Select, which uses pairs of problems and their corresponding generated reasoning paths to automatically construct effective demonstrations.
Specifically, it defines the confusion score of each problem in the training set: the score is higher, if the semantic relevance of its conditions is lower; and the problems with the highest confusion scores are selected.

Experiments on GPT-3.5-Turbo demonstrate that adding the I$^3$C instruction to CoT prompting methods improves their performance.
For example, adding I$^3$C instruction to Manual-CoT improves the accuracy by $+8.1$ on AddSub, $+8.1$ on SVAMP, $+6.0$ on GSM8K, $+5.1$ on SingleEq, $+5.1$ on GSM-IC2-1K, $+2.8$ on AQuA, $+9.2$ on MATH, and $+7.8$ on GSM-ICM-1K.

Moreover, I$^3$C-Select beats existing prompting methods by a striking margin on eight MWP datasets.
Specifically, I$^3$C-Select boosts the performance of Complex-CoT method by $+11.7$ on GSM-IC2-1K, $+11.1$ on GSM-ICM-1K, $+12.6$ on AQuA, $+8.2$ on MATH, and $+10.0$ on GSM8K.

\section{Related Work}
\label{sec:related}
\subsection{Math Word Problem Solving}
Our work is related to existing efforts on solving MWPs.
Traditional methods used statistical learning to extract entities, quantities, and operators from a question and generated an arithmetic equation to find the answer \citep{Hosseini-2014-AddSub,roy-2015-singleop,zhou-2015-qp,mitra-2016-mathstudent,hu2022heterogeneous}.
Later, sequence-to-sequence (Seq2Seq) model and recurrent neural networks directly transformed the question into an arithmetic equation \citep{wang-2017-dns,wang-2019-template, li-2019-modeling}. 
Recently, fine-tuned pre-trained language models have significantly improved the validity of generated equations and accuracy of answers \citep{shen-2021-generate-rank,liang-2023-wda}.
However, these methods require a large amount of human annotations, lacking the ability to generalize to new kinds of MWPs.
In this work, we aim to prompt LLMs to answer arbitrary MWPs and generate reasoning paths, without human annotations or task-specific fine-tuning.

\subsection{Chain-of-Thought Prompting Methods}
CoT prompting methods have enabled LLMs to generate reasoning paths and solve complex MWPs \citep{Kojima-2022-CoT}. The reasoning paths could be more expressive if the prompts were added with ``\emph{Let's think step by step}''. To mitigate missing-step errors, Plan-and-Solve (PS) prompting methods instructed the LLMs to devise a plan to break down the entire task into smaller subtasks, and then carry out the subtasks according to the plan \citep{wang-2023-ps}.
Manual-CoT, as a type of few-shot prompting \cite{ziems2023large}, manually designed demonstrations to elicit multi-step reasoning ability of the LLMs \citep{wei-2022-cot}.
Program of Thought (PoT) generated programming language statements and used a program interpreter to execute the generated program to get final answers \citep{chen-2022-pot}.
\citeauthor{zhang-2023-autocot} designed Auto-CoT, and their source code\footnote{\url{https://github.com/amazon-science/auto-cot}} showed that they sampled diverse questions from the test set to minimize manual effort in finding demonstrations.
\citeauthor{fu-2023-complexity} designed Complex-CoT, which selects the most complex problems and their reasoning paths as demonstrations.
Aware of irrelevant conditions in the problem description, \citeauthor{freda-2023-gsm8k-ic} added the instruction ``\emph{Feel free to ignore irrelevant conditions in the problem description}'' in the prompt. 
These methods do not explicitly specify the irrelevant conditions in the prompt, which makes it difficult for LLMs to identify and ignore irrelevant conditions in the problem solving process.
Our method identifies irrelevant conditions in the problem description, instructs the LLMs to ignore them, and achieves significantly higher accuracy.

\subsection{Identify Irrelevant Information}
\citeauthor{jia-liang-2017-adversarial} have shown that question answering systems are confused when paragraphs contain irrelevant information.
Several studies have trained models to identify and filter out the irrelevant information.
For example, \citeauthor{roy-roth-2015-solving} trained a classifier and scored the likelihood of each quantity in the problem being an irrelevant quantity.
\citeauthor{Junhan-2022-GZSL} employed a new training loss to remove the attribute-irrelevant information from the semantic encoder output.
\citeauthor{Li-2022-MsKAT} proposed a multi-scale knowledge-aware transformer to eliminate identity-irrelevant information.
\citeauthor{yang-2023-img} leveraged pre-extracted semantic information to improve the preprocessor's ability to accurately identify and filter out task-irrelevant information.
All these methods require massive human annotations.
In contrast, our method does not require time-consuming training or fine-tuning. It employs LLMs to automatically identify irrelevant conditions and generate instructions to help the models ignore them.

\section{Proposed Approach}
\label{sec:method}
\subsection{Overview}
In this section, we elaborate on how to instruct LLMs to identify and ignore irrelevant conditions in the math word problem description.
Given a complex problem, we first identify a set of irrelevant condition candidates that have a weak semantic relevance with the question (\cref{sec:identify}).
Then we prompt LLMs to verify if the candidates are indeed irrelevant.
Putting all the verification results together, we create a novel I$^3$C instruction to instruct the LLMs to ignore the irrelevant conditions in the problem description.
The I$^3$C instruction can be added to any CoT prompting methods to help LLMs avoid confusion and improve their generated reasoning paths.
Furthermore, we develop a few-shot prompting method I$^3$C-Select that selects the most confusing problems and their reasoning paths as demonstrations, and adds the I$^3$C instruction before the demonstrations in the prompt.
Given the prompt and a target problem, the LLMs generate an accurate reasoning path to improve the solving process.
We introduce the I$^3$C instruction in \cref{sec:i3c_instruction} and I$^3$C-Select method in \cref{sec:i3c_select}.

\subsection{Identify a Set of Irrelevant Condition Candidates}
\label{sec:identify}
Given a MWP $Q$, we first split it into $n$ conditions $\{c_{i}\}_{i=1}^{n}$ and a question sentence $q$, where each condition describes at most one quantity. So we have $Q=[\{c_i\},q]$.
Specifically, for a MWP, we initially segment it into multiple sentences based on the full stop. Then, we select the last sentence as the question sentence. For the remaining sentences, we further analysis whether they contain multiple quantities and commas. If they do, we segment the sentence into multiple sub-sentences based on the commas, with each sub-sentence containing at most one quantity.
For example, in Figure~\ref{fig:framework_1}, the conditions are \{``\emph{Steve is 5'6".}'', ``\emph{He grows 6 inches.}'', ``\emph{The height of Mary is 30 feet.}''\}, and the question sentence is ``\emph{How tall is Steve in inches?}''.

Next, we use a pre-trained language model, e.g., SimCSE~\citep{gao-2021-simcse}, to encode the conditions and question sentence into vector representations. So we have $\{\mathbf{c}_{i}\}_{i=1}^{n}$ and $\mathbf{q}$ which are $d$-dimensional vectors. We set $d=1,024$.

Then for each condition $c_i$, we calculate the average similarity between $c_i$ and all other conditions in $Q$ using cosine similarity, because the SimCSE embeddings were trained on cosine similarity:
\begin{equation}
\begin{aligned}
    s_{i}^{\text{(c)}} &= \frac{1}{n-1} \sum_{j=1, j \neq i}^n \cos{(\mathbf{c}_i, \mathbf{c}_j)} \\ &= \frac{1}{n-1} \sum_{j=1, j \neq i}^n \frac{\mathbf{c}_i^{\top} \mathbf{c}_j}{{\|\mathbf{c}_i\|}\cdot{\|\mathbf{c}_j\|}}.
\end{aligned}
\end{equation}
We also calculate the similarity between $c_i$ and $q$: $s_{i}^{\text{(q)}} = \cos{(\mathbf{c}_i, \mathbf{q})}$.
So we have $\{s_{i}^{\text{(c)}}, s_{i}^{\text{(q)}}\}_{i=1}^{n}$.

Now we can define a set of \emph{irrelevant condition candidates} $\mathcal{I} \subset \{c_{i}\}_{i=1}^{n}$ for each math word problem. A condition $c_i$ is potentially irrelevant if its semantic relevance is lower than a threshold. In other words, if $s_{i}^{\text{(c)}} < \theta$ or $s_{i}^{\text{(q)}} < \theta$, $\mathcal{I}$ has $c_i$. We re-index the conditions in the set: $\mathcal{I}=\{c_{k}^{(\mathrm{irr})}\}_{k=1}^{|\mathcal{I}|}$. The threshold $\theta$ is a hyperparameter. We set $\theta=0.5$.
See Appendix~\ref{sec:appendix_results} for hyperparameter analysis.

We can further define the \emph{confusion score} of a math word problem $Q$. We assume that the problem is more confusing if its conditions are less relevant with the final question. So the confusion score is defined as the inverse of the average similarity between any condition and the question:
\begin{equation}
    \text{conf}(Q) = \left[\frac{1}{n}\sum_{i=1}^n\cos{(\mathbf{c}_i, \mathbf{q})}\right]^{-1}.
    \label{eq:confusion_score}
\end{equation}
The most confusing problems, i.e., the problems of the highest confusion scores, and their generated reasoning paths, will be automatically used as demonstrations in a few-shot setting. The demos teach LLMs to better solve confusing problems. Later sections give details.

\subsection{Construct I$^3$C Instruction}
\label{sec:i3c_instruction}

Given a set of irrelevant condition candidates $\mathcal{I}$, we use an LLM to verify if the candidates are indeed irrelevant.
For a math word problem $Q$, its final question $q$, and a condition candidate $c_{k}^{(\mathrm{irr})} \in \mathcal{I}$, we construct a verification prompt: ``\emph{$Q$. Is condition $c_{k}^{(\mathrm{irr})}$ relevant to the process of solving problem $q$?}''
We feed the prompt to an LLM and receive a piece of text $v_{k}^{(\mathrm{irr})}$ justifying if $c_{k}^{(\mathrm{irr})}$ is relevant or indeed irrelevant.
So we have a set of verification outputs (size $|\mathcal{I}|$): $\{v_{k}^{(\mathrm{irr})}\}_{k=1}^{|\mathcal{I}|}$.

Now we can create a novel instruction to help LLMs identify and ignore irrelevant conditions in the problem description. In a zero-shot setting, the instruction starts with all the verification outputs.
Specifically, this I$^3$C instruction, simply denoted by $I$, is ``\emph{The instructions are as follows: $v_{1}^{(\mathrm{irr})}$ $\cdots$ $v_{|\mathcal{I}|}^{(\mathrm{irr})}$. Let's consider these instructions and ignore the irrelevant conditions to solve the problem}''.
In case where $\mathcal{I}$ is an empty set, we follow the Instruct-CoT method~\citep{freda-2023-gsm8k-ic} and use the sentence ``\emph{Feel free to ignore irrelevant conditions in the problem description}'' as the instruction.

\subsection{Generate Reasoning Paths and Answers with I$^3$C Instruction}
The I$^3$C instruction can be added to any CoT prompting methods such as Zero-Shot-CoT~\citep{Kojima-2022-CoT}, PS~\citep{wang-2023-ps}, Instruct-CoT~\citep{freda-2023-gsm8k-ic}, Manual-CoT~\citep{wei-2022-cot}, Complex-CoT~\citep{fu-2023-complexity}, and Auto-CoT~\citep{zhang-2023-autocot}. The goal is to generate a reasoning path and answer a math word problem $Q$.
For example, in Zero-Shot-CoT~\citep{Kojima-2022-CoT}, the prompt was ``\emph{Q: $Q$. A: Let's think step by step}''.
By adding the I$^3$C instruction to the Zero-Shot-CoT method, denoted by Zero-Shot-CoT+I$^3$C in our experiments, the prompt becomes ``\emph{\underline{$I$.} Q: $Q$. A: Let's think step by step}''.
The full prompts in experiments can be found in Appendix~\ref{sec:appendix_prompts}.

Finally, after the reasoning path is generated, we use the prompt ``\emph{Therefore, the answer is}'' to get the quantity prediction as the final answer.

\subsection{I$^3$C-Select: Select Confusing Problems as Automatic Demonstrations}
\label{sec:i3c_select}
\citeauthor{fu-2023-complexity} found that prompts with higher reasoning complexity achieved better performance on multi-step reasoning tasks.
To further enhance the ability of LLMs to address the irrelevance of conditions, we develop a novel few-shot prompting method I$^3$C-Select.
As presented in \cref{sec:identify}, it first calculates the confusion score of problems in the training set, as defined in Eq.(\ref{eq:confusion_score}).
Subsequently, it selects the $K$ most confusing problems and generates their reasoning paths using the Zero-Shot-CoT prompting method (with $K=8$ in our experiments).
Finally, it uses the most confusing problems and their reasoning paths as demonstrations, denoted by $\{Q_{1}^{\text{(demo)}},R_{1}^{\text{(demo)}};\cdots;Q_{K}^{\text{(demo)}},R_{K}^{\text{(demo)}}\}$.

I$^3$C-Select puts the demonstrations after the I$^3$C instruction to construct the full prompt. Specifically, the prompt is ``\emph{\underline{$I$.} Q: $Q_{1}^{\emph{\text{(demo)}}}$  A: $R_{1}^{\emph{\text{(demo)}}}$ $\cdots$ Q: $Q_{K}^{\emph{\text{(demo)}}}$ A: $R_{K}^{\emph{\text{(demo)}}}$ Q: $Q$. A:}''.
With the prompt and the target problem $Q$, the LLMs generate a reasoning path for $Q$. Figure~\ref{fig:framework_2} illustrates the details.

\section{Experiments}
\label{sec:experiments}
\begin{table*}[t]
\renewcommand\arraystretch{1.15}
\caption{Accuracy (\%) comparison on eight MWP datasets. I$^3$C indicates that instructs LLMs to identify and ignore irrelevant conditions. Adding the I$^3$C instruction to CoT prompting methods effectively improves performance. Selecting the most confusing problems and their generated reasoning paths as demonstrations for few-shot learning (i.e., I$^3$C-Select) achieves state-of-the-art performance on all eight MWP datasets.}
\centering
\begin{adjustbox}{width=\textwidth,center}
\begin{tabular}{l|l|p{1.6cm}p{1.6cm}p{1.7cm}p{1.6cm}p{1.6cm}p{1.6cm}p{1.7cm}p{1.6cm}}
\bottomrule
 \multicolumn{1}{l|}{\multirow{2}{*}{\makecell[c]{LLM}}} & \multicolumn{1}{l|}{\multirow{2}{*}{\makecell[c]{Method}}} & \multicolumn{8}{c}{Dataset} 
 \\ \cline{3-10}
 & \multicolumn{1}{c|}{} & \multicolumn{1}{l}{AddSub} & \multicolumn{1}{l}{SVAMP} & \multicolumn{1}{l}{GSM8K} & \multicolumn{1}{l}{SingleEq} & \multicolumn{1}{l}{GSM-IC2-1K} & \multicolumn{1}{l}{GSM-ICM-1K} & \multicolumn{1}{l}{AQuA} & \multicolumn{1}{l}{MATH}\\ 
 \hline
 \multicolumn{1}{l|}{\multirow{15}*{\rotatebox{90}{\makecell[c]{GPT-3 (text-davinci-003)}}}}
 & Direct & $89.3$ & $65.2$ & $15.0$ & $84.6$ & $22.8$ & $9.0$ & $28.7$ & $7.6$ \\
 & Direct + I$^3$C & $92.4  \ (+3.1)$ & $74.5 \ (+9.3)$ & $49.7  \ (+34.7)$ & $92.7  \ (+8.1)$ & $82.6 \ (+59.8)$ & $66.9\ (+57.9)$ & $36.2 \ (+7.5)$ & $11.3 \ (+3.7)$ \\
 \cline{2-10}
 
 & Zero-Shot-CoT & $84.8$ & $74.3$ & $60.8$ & $89.5$ & $70.7$ & $62.5$ & $40.5$ & $12.4$ \\
 & Zero-Shot-CoT + I$^3$C & $91.7\ (+6.9)$ & $75.9\ (+1.6)$ & $61.3  \ (+0.5)$ & $93.7 \ (+4.2)$ & $84.7 \ (+14.0)$ & $71.4  \ (+8.9)$ & $45.7 \ (+5.2)$ & $17.9 \ (+5.5)$ \\
 \cline{2-10}
 
 & PS & $88.1$ & $72.0$ & $58.2$ & $89.2$ & $70.9$ & $63.5$ & $38.1$ & $13.7$ \\
 & PS + I$^3$C & $91.4\ (+3.3)$ & $75.6\ (+3.6)$ & $61.1  \ (+2.9)$ & $93.1\ (+3.9)$ & $84.8 \ (+13.9)$ & $69.4  \ (+5.9)$ & $43.6 \ (+5.5)$ & $18.2 \ (+4.5)$ \\
 \cline{2-10}
 
 & Instruct-CoT & $90.4$ & $76.3$ & $57.8$ & $91.1$ & $82.4$ & $64.3$ & $44.5$ & $16.1$ \\
 & Instruct-CoT + I$^3$C & $91.8\ (+1.4)$ & $77.0 \ (+0.7)$ & $61.0  \ (+3.2)$ & $92.7\ (+1.6)$ & $84.7 \ (+2.3)$ & $71.3\ (+7.0)$ & $46.3 \ (+1.8)$ & $21.3 \ (+5.2)$ \\
 
 \cline{2-10}
 & Manual-CoT & $87.8$ & $76.7$ & $56.9$ & $91.3$ & $73.9$ & $60.6$ & $44.0$ & $15.6$ \\
 & Manual-CoT + I$^3$C & $92.9 \ (+5.1)$ & $80.1  \ (+3.4)$ & $61.6 \ (+4.7)$ & $93.9  \ (+2.6)$ & $82.0 \ (+8.1)$ & $66.1\ (+5.5)$ & $49.1 \ (+5.1)$ & $19.8 \ (+4.2)$ \\
 \cline{2-10}
 
 & Auto-CoT & $90.6$ & $77.8$ & $58.9$ & $90.9$ & $74.3$ & $65.2$ & $47.2$ & $16.3$ \\ 
 & Auto-CoT + I$^3$C & $93.7 \ (+3.1)$ & $80.0  \ (+2.2)$ & $61.9  \ (+3.0)$ & $93.5  \ (+2.6)$ & $83.9 \ (+9.6)$ & $68.2  \ (+3.0)$ & $51.5 \ (+4.3)$ & $22.5 \ (+6.2)$ \\
 \cline{2-10}

 & Complex-CoT & $88.9$ & $78.0$ & $67.7$ & $92.7$ & $75.3$ & $66.5$ & $48.8$ & $17.4$ \\ 
 & Complex-CoT + I$^3$C & $92.8 \ (+3.9)$ & $80.0 \ (+2.0)$ & $70.6 \ (+2.9)$ & $94.0 \ (+1.3)$ & $87.1 \ (+11.8)$ & $83.6 \ (+17.1)$  & $53.2 \ (+4.4)$  & $23.1 \ (+5.7)$ \\
 \cline{2-10}
 
 & I$^3$C-Select (Ours) & $\mathbf{93.9}$ & $\mathbf{80.3}$ & $\mathbf{72.6}$ & $\mathbf{94.3}$ & $\mathbf{93.7}$ & $\mathbf{90.9}$ & $\mathbf{57.1}$ & $\mathbf{28.5}$ \\
\hline
 \multicolumn{1}{l|}{\multirow{16}*{\rotatebox{90}{\makecell[c]{GPT-3.5-Turbo-1106}}}}
 & Direct & $86.1$ & $78.2$ & $77.8$ & $93.1$ & $88.9$ & $83.4$ & $63.4$ & $39.7$ \\ 
 & Direct + I$^3$C & $94.4 \ (+8.3)$ & $85.1 \ (+6.9)$ & $78.5 \ (+0.7)$ & $96.9 \ (+3.8)$ & $92.5 \ (+3.6)$ & $90.1 \ (+6.7)$  & $64.2 \ (+0.8)$  & $41.3 \ (+1.6)$ \\
 \cline{2-10}
 
 & Zero-Shot-CoT & $85.2$ & $76.7$ & $78.6$ & $90.3$ & $87.0$ & $82.0$ & $51.3$ & $37.9$ \\ 
 & Zero-Shot-CoT + I$^3$C & $93.4 \ (+8.2)$ & $84.2 \ (+7.5)$ & $82.0 \ (+3.4)$ & $97.8 \ (+7.5)$ & $92.7 \ (+5.7)$ & $88.6 \ (+6.6)$  & $63.1 \ (+11.8)$  & $42.1 \ (+4.2)$ \\
 \cline{2-10}
 
 & PS & $87.6$ & $77.8$ & $75.9$ & $91.7$ & $81.4$ & $73.6$ & $60.2$ & $43.7$ \\ 
 & PS + I$^3$C & $93.7 \ (+6.1)$ & $85.6 \ (+7.8)$ & $82.5 \ (+6.6)$ & $97.6 \ (+5.9)$ & $92.7 \ (+11.3)$ & $90.1 \ (+16.5)$  & $64.5 \ (+4.3)$  & $45.2 \ (+1.5)$ \\
 \cline{2-10}
 
 & Instruct-CoT & $86.5$ & $81.3$ & $77.7$ & $94.4$ & $89.2$ & $84.4$ & $62.9$ & $41.1$ \\ 
 & Instruct-CoT + I$^3$C & $92.9 \ (+6.4)$ & $84.9 \ (+3.6)$ & $82.0 \ (+4.3)$ & $97.8 \ (+3.4)$ & $92.9 \ (+3.7)$ & $89.1 \ (+4.7)$  & $65.5 \ (+2.6)$  & $46.1 \ (+5.0)$ \\
 
 \cline{2-10}
 & Manual-CoT & $85.3$ & $77.1$ & $76.4$ & $92.9$ & $86.8$ & $81.4$ & $54.3$ & $35.1$ \\ 
 & Manual-CoT + I$^3$C & $93.4 \ (+8.1)$ & $85.2 \ (+8.1)$ & $82.4 \ (+6.0)$ & $98.0 \ (+5.1)$ & $91.9 \ (+5.1)$ & $89.2 \ (+7.8)$  & $57.1 \ (+2.8)$  & $44.3 \ (+9.2)$ \\
 \cline{2-10}
 
 & Auto-CoT & $88.0$ & $80.9$ & $78.8$ & $95.9$ & $84.3$ & $81.8$ & $57.8$ & $39.1$ \\ 
 & Auto-CoT + I$^3$C & $93.2 \ (+5.2)$ & $84.7 \ (+3.8)$ & $82.8 \ (+4.0)$ & $97.8 \ (+1.9)$ & $91.8 \ (+7.5)$ & $88.4 \ (+6.6)$  & $62.7 \ (+4.9)$  & $43.9 \ (+4.8)$ \\
 \cline{2-10}

 & Complex-CoT & $87.9$ & $80.4$ & $78.9$ & $94.5$ & $84.3$ & $83.0$ & $59.1$ & $39.5$ \\ 
 & Complex-CoT + I$^3$C & $93.7 \ (+5.8)$ & $84.4 \ (+4.0)$ & $82.4 \ (+3.5)$ & $97.2 \ (+2.7)$ & $91.7 \ (+7.4)$ & $88.6 \ (+5.6)$  & $63.2 \ (+4.1)$  & $45.3 \ (+5.8)$ \\
 \cline{2-10}

 & PAL & $89.1$ & $77.8$ & $79.5$ & $97.6$ & $85.2$ & $84.7$ & $63.4$ & $38.7$ \\ 
 \cline{2-10}
 
 & I$^3$C-Select (Ours) & $\mathbf{94.9}$ & $\mathbf{89.9}$ & $\mathbf{88.9}$ & $\mathbf{98.6}$ & $\mathbf{96.0}$ & $\mathbf{94.1}$ & $\mathbf{71.7}$ & $\mathbf{47.7}$ \\

\toprule
\end{tabular}
\end{adjustbox}
\label{tab:mainresult}
\end{table*}

\subsection{Experimental Setup}
\paragraph{Datasets.} 
We use eight math word problem (MWP) datasets as our testbed.
AddSub~\citep{Hosseini-2014-AddSub}, SingleEq~\citep{koncel-2015-singleq}, SVAMP~\citep{patel-2021-svamp}, and GSM8K~\citep{karl-2021-gsm8k} are classical MWP datasets in which some of the problem descriptions contain irrelevant conditions.
GSM-IC2-1K~\citep{freda-2023-gsm8k-ic} and GSM-ICM-1K~\citep{freda-2023-gsm8k-ic} are challenging datasets that require multi-step reasoning,
and each problem description contains irrelevant conditions.
AQuA~\citep{ling-etal-2017-aqua} and MATH~\citep{dan-etal-2021-MATH} are more challenging datasets that contain problems from high school competitions.
More detailed dataset information can be found in Appendix~\ref{sec:appendix_datasets}.

\paragraph{Baselines.} 
We compare our proposed I$^3$C-Select prompting method with two types of prompting baselines: 
(1) Zero-shot baselines. We include Zero-Shot-CoT~\citep{Kojima-2022-CoT}, PS~\citep{wang-2023-ps}, Instruct-CoT~\citep{freda-2023-gsm8k-ic}, and Direct~\citep{Kojima-2022-CoT}.
The Direct baseline uses the prompt ``\emph{The answer is}'' to get the final answer.
(2) Few-shot baselines. We include Manual-CoT~\citep{wei-2022-cot}, Complex-CoT~\citep{fu-2023-complexity}, PAL~\citep{Gao-2023-pal}, and Auto-CoT~\citep{zhang-2023-autocot}. 
The demonstrations of these baselines are from their original papers.
Notably, according to the source code, Auto-CoT's demonstrations are from the test set, whereas I$^3$C-Select's demonstrations are from the training set.
Details of all baselines are shown in Appendix~\ref{sec:appendix_baselines}.

\paragraph{Implementation.} 
We use GPT-3 (text-davinci-003) and GPT-3.5-Turbo-1106 as backend LLMs, which are the most widely-used LLMs with public APIs.
Following~\citep{freda-2023-gsm8k-ic}, we set the temperature to $0.7$. 
To evaluate the model performance, we follow~\citep{chen-2022-pot} to adopt accuracy as our evaluation metric.
An answer is considered correct if and only if the absolute error between the answer and the gold answer is less than $1 \times 10^{-5}$.
See Appendix~\ref{sec:appendix_metrics} for detail.

\subsection{Experimental Results}
\paragraph{Overall performance on MWP datasets.}
As shown in Table~\ref{tab:mainresult}, I$^3$C-Select consistently outperforms the baseline methods across all MWP datasets by a significant margin, regardless of which model is used as the backend LLM.
Specifically, when applied to GPT-3 (text-davinci-003), I$^3$C-Select improves the accuracy over Zero-Shot-CoT by at least $+6.0$ across all datasets, except for SingleEq, where the improvement is $+4.8$.  
This exception can be attributed to the fact that the problems in SingleEq do not contain irrelevant conditions. 
Our proposed I$^3$C-Select method primarily instructs LLMs to identify and ignore irrelevant conditions in the problem description. 
It is noteworthy that even in the SingleEq dataset, using the most confusing problems and their generated reasoning paths as demonstrations effectively enhances MWP solving performance. 

In comparison to the competitive zero-shot baseline, Instruct-CoT, the performance of I$^3$C-Select remains impressive. 
When applied to GPT-3.5-Turbo, I$^3$C-Select enhances the average accuracy by $+8.0$ across eight MWP datasets compared to Instruct-CoT. 
Furthermore, our analysis demonstrates that I$^3$C-Select consistently outperforms few-shot baselines on all datasets. 
Specifically, when compared to the Complex-CoT prompting method, I$^3$C-Select exhibits superior performance in GSM-ICM-1K, GSM-IC2-1K, AQuA, MATH, and GSM8K, with improvements of $+11.1$, $+11.7$, $+12.6$, $+8.2$, and $+10.0$, respectively. 
These findings indicate that incorporating more detailed instructions (e.g., I$^3$C instruction) and using the most confusing problems and their reasoning paths in the prompt can achieve superior performance.

\begin{table*}[t]
\renewcommand\arraystretch{1.15}
\caption{Accuracy (\%) on GSM-IC-2K dataset, broken down by the number of reasoning steps required in the standard answer. The GSM-IC-2K dataset is formed by merging the GSM-IC2-1K and GSM-ICM-1K datasets.}
\centering
\begin{adjustbox}{width=0.75\textwidth,center}
\begin{tabular}{l|p{1.75cm}p{1.75cm}p{1.75cm}p{1.75cm}p{1.75cm}}
\bottomrule
 \multicolumn{1}{l|}{\multirow{2}{*}{\makecell{Method\\(GPT-3.5-Turbo)}}} & \multicolumn{5}{c}{Accuracy by Steps (GSM-IC-2K)} 
 \\ \cline{2-6}
 \multicolumn{1}{c|}{} & \multicolumn{1}{l}{2 Steps} & \multicolumn{1}{l}{3 Steps} & \multicolumn{1}{l}{4 Steps} & \multicolumn{1}{l}{$\geq$ 5 Steps} & \multicolumn{1}{l}{All} \\ 
 \hline
 
 Zero-Shot-CoT & $87.0$ & $82.0$ & $80.2$ & $82.6$ & $84.5$ \\
 Zero-Shot-CoT + I$^3$C & $92.7 \ (+5.7)$ & $91.4 \ (+9.4)$ & $81.3 \ (+1.1)$ & $92.4 \ (+9.8)$ & $90.7 \ (+6.2)$ \\
 \hline
 
 
 Instruct-CoT & $89.2$ & $85.8$ & $81.3$ & $84.6$ & $86.8$ \\
 Instruct-CoT + I$^3$C & $92.9 \ (+3.7)$ & $90.6 \ (+4.8)$ & $82.3 \ (+1.0)$ & $93.9 \ (+9.3)$ & $91.0 \ (+4.2)$ \\
 
 \hline
 Manual-CoT & $86.8$ & $85.0$ & $78.8$ & $79.7$ & $84.1$ \\
 Manual-CoT + I$^3$C & $91.9 \ (+5.1)$ & $90.6 \ (+5.6)$ & $80.6 \ (+1.8)$ & $94.8 \ (+15.1)$ & $90.6 \ (+6.5)$ \\
 \hline
 

 Complex-CoT & $84.3$ & $81.0$ & $83.4$ & $84.6$ & $83.7$ \\
 Complex-CoT + I$^3$C & $91.7 \ (+7.4)$ & $89.8 \ (+8.8)$ & $83.8 \ (+0.4)$ & $91.6 \ (+7.0)$ & $90.2 \ (+6.5)$ \\
 \hline

 I$^3$C-Select (Ours) & $\mathbf{96.0}$ & $\mathbf{95.2}$ & $\mathbf{87.3}$ & $\mathbf{98.6}$ & $\mathbf{95.1}$ \\
\toprule
\end{tabular}
\end{adjustbox}
\label{tab:stepsresult}
\end{table*}

\begin{figure*}[t]
  \centering
  \includegraphics[width=0.94\textwidth]{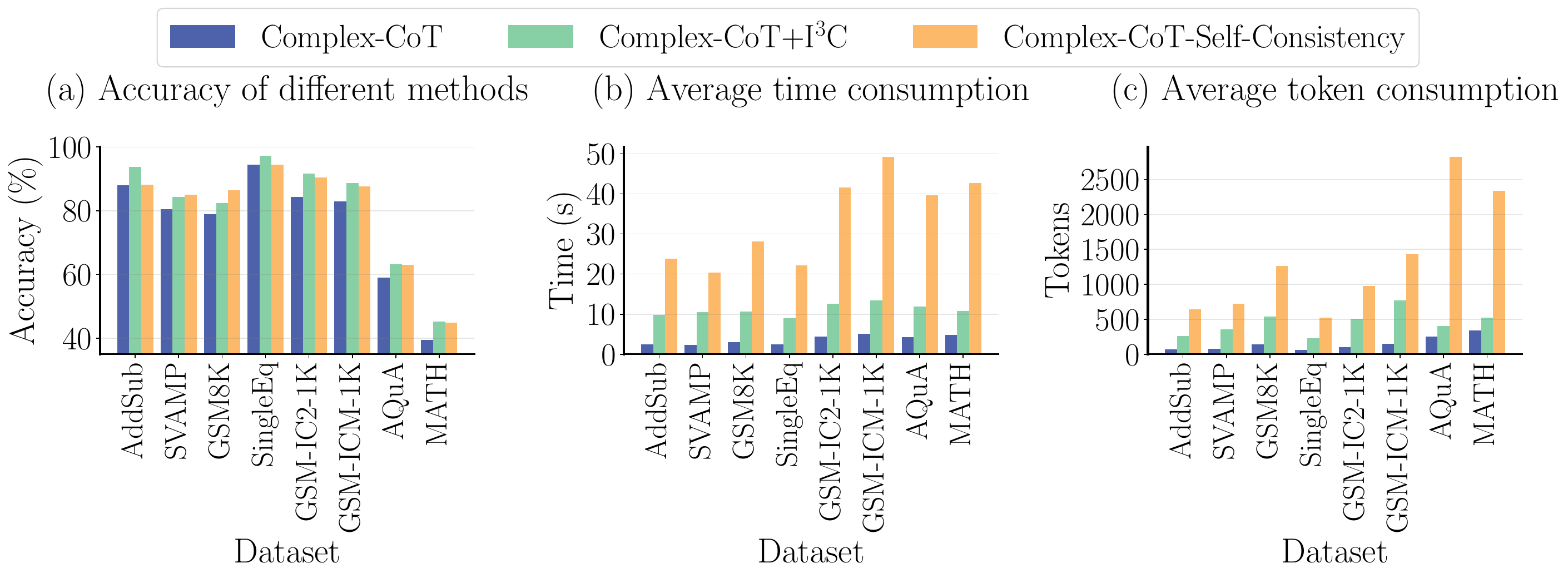}
  \caption{Performance comparison of Complex-CoT, Complex-CoT with I$^3$C instruction (i.e., Complex-CoT+I$^3$C), and Complex-CoT with self-consistency (i.e., Complex-CoT-Self-Consistency). We can observe that the accuracy of Complex-CoT+I$^3$C and Complex-CoT-Self-Consistency is nearly identical, while Complex-CoT+I$^3$C consumes much less tokens and time than Complex-CoT-Self-Consistency.}
  \label{fig:I3C_SC}
\end{figure*}

\paragraph{Does adding the I$^3$C instruction work?}
As shown in Table~\ref{tab:mainresult}, adding the I$^3$C instruction to the CoT prompting methods significantly enhances the MWP solving performance.
Specifically, when applied to GPT-3.5-Turbo, adding the I$^3$C instruction to the Zero-Shot-CoT method (i.e., Zero-Shot-CoT+I$^3$C) improves the average accuracy by $+6.9$ across eight MWP datasets, compared to the original Zero-Shot-CoT prompting method. 
For datasets like GSM-IC2-1K and GSM-ICM-1K, which contain irrelevant conditions in each problem description, Zero-Shot-CoT+I$^3$C improves the accuracy by $+5.7$ and $+6.6$, respectively.
Even for prompting methods such as Auto-CoT, which already achieve high accuracy on most MWP datasets, the addition of the I$^3$C instruction (i.e., Auto-CoT+I$^3$C) still leads to significant improvements.
Auto-CoT+I$^3$C improves accuracy by $+7.5$ on GSM-IC2-1K, $+4.9$ on AQuA, $+4.8$ on MATH, and $+4.0$ on GSM8K.

\paragraph{How does LLM selection affect I$^3$C-Select?}
Table~\ref{tab:mainresult} shows that I$^3$C-Select works better when the LLM is more powerful.
Specifically, on the GSM8K dataset, the GPT-3.5-Turbo model exhibits a $+16.3$ increase in accuracy compared to the text-davinci-003 model.
Similarly, on the AQuA dataset, using the GPT-3.5-Turbo model results in a $+14.6$ improvement in accuracy over the text-davinci-003 model. 
It is noteworthy that GPT-3.5-Turbo is a chat-optimized model built upon text-davinci-003 \citep{zheng-2023-gptfathom}. 
The enhanced performance with GPT-3.5-Turbo can be attributed to its enhanced power, making it better at understanding and utilizing the given prompt.

\paragraph{Compared with executor-augmented prompting methods.}
Table~\ref{tab:mainresult} shows that I$^3$C-Select consistently outperforms the executor-augmented prompting methods, such as PAL, across all MWP datasets. 
Specifically, in comparison to the PAL prompting method, I$^3$C-Select exhibits superior performance in GSM-IC2-1K, AQuA, SVAMP, AddSub, and GSM8K, with improvements of $+10.8$, $+8.3$, $+12.1$, $+5.8$, and $+9.4$, respectively.


\paragraph{Does I$^3$C instruction work for complex problems?}
We analyze the breakdown accuracies for problems with respect to the reasoning steps\footnote{Number of reasoning steps of a problem is given by the number of sentences in standard answer.~\citep{karl-2021-gsm8k}} in Table~\ref{tab:stepsresult}.
The GSM-IC-2K dataset is formed by merging the GSM-IC2-1K and GSM-ICM-1K datasets.
Each problem in GSM-IC-2K contains irrelevant conditions and requires multiple steps to solve.
Obviously, adding the I$^3$C instruction to the CoT prompting method significantly enhances the MWP solution performance for both simple and complex problems.
Moreover, compared to Complex-CoT, I$^3$C-Select significantly improves the performance on GSM-IC-2K: from $83.7$ to $95.1$. 
These results indicate that adding the I$^3$C instruction to the prompt can effectively solve complex problems.

\paragraph{Efficiency and effectiveness of I$^3$C instruction.}
Self-consistency~\citep{wang-2023-selfconsistency} is the process of solving a problem $M$ times and using a majority vote strategy to determine the most consistent answer as the final answer.
We evaluate the performance of Complex-CoT with self-consistency (i.e., Complex-CoT-Self-Consistency) on eight MWP datasets.
Following~\citep{wang-2023-ps}, we set $M$ to $10$.
Figure~\ref{fig:I3C_SC} shows that the accuracy of Complex-CoT-Self-Consistency and Complex-CoT+I$^3$C is nearly identical.
In terms of time consumption\footnote{Efficiency analysis for Complex-CoT+I$^3$C considers the cost of (1) running SimCSE for each problem, (2) using  LLM as a verifier, and (3) prompting LLM to solve the problem.}, Complex-CoT+I$^3$C proves to be an efficient method, reducing the average time required to solve an MWP by $2$-$4$ times compared to Complex-CoT-Self-Consistency.
Regarding token consumption, Complex-CoT+I$^3$C consumes fewer tokens than Complex-CoT-Self-Consistency, indicating its more concise and efficient nature in solving MWPs.
Overall, the results demonstrate that Complex-CoT+I$^3$C consumes much fewer computational resources than Complex-CoT-Self-Consistency while maintaining comparable accuracy.

\begin{figure}[t]
  \centering
  \includegraphics[width=1.0\columnwidth]{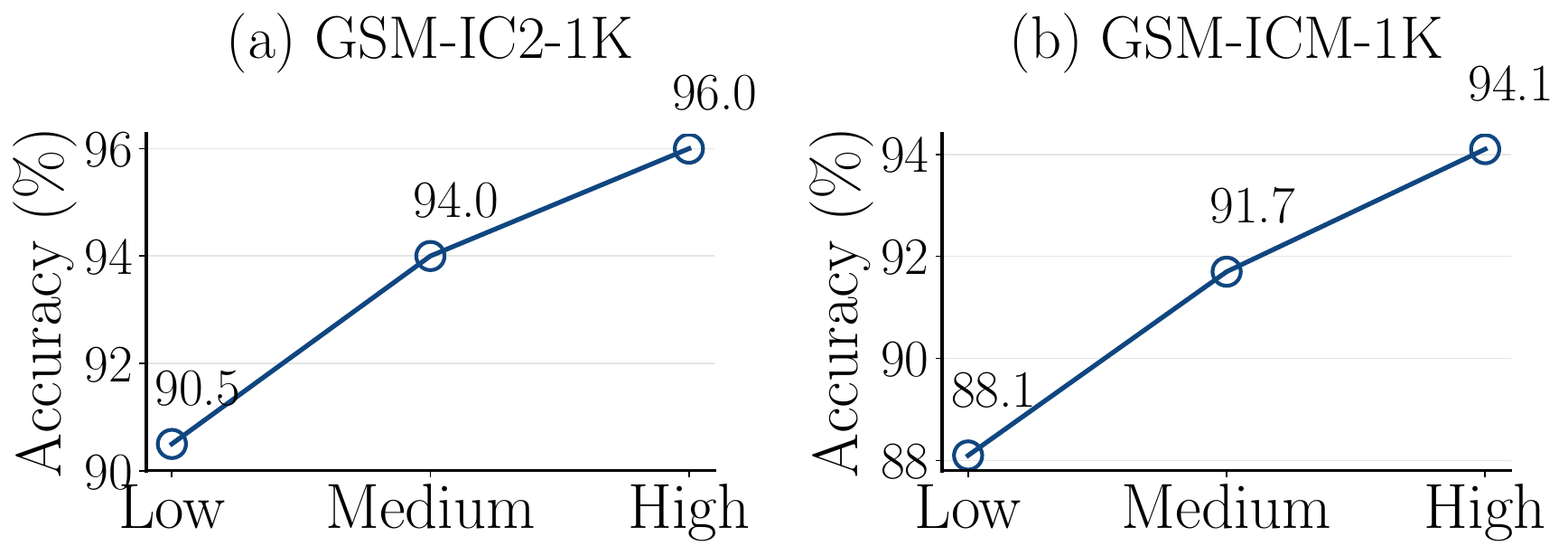}
  \caption{Demonstration construction methods comparison. ``Low'' indicates selecting eight problems with the lowest confusion scores. ``Medium'' indicates randomly selecting eight problems. ``High'' indicates selecting eight problems with the highest confusion scores.}
  \label{fig:easy-medium}
\end{figure}

\begin{table}[t]
\renewcommand\arraystretch{1.15}
\caption{Accuracy (\%) comparison of different methods that help LLMs ignore irrelevant conditions.}
\centering
\begin{adjustbox}{width=1.0\columnwidth,center}
\begin{tabular}{l|p{1.6cm}<{\centering}p{1.6cm}<{\centering}}
\bottomrule
 \multicolumn{1}{l|}{\multirow{2}{*}{\makecell{Method\\(GPT-3.5-Turbo)}}} & \multicolumn{2}{c}{Dataset} 
 \\ \cline{2-3}
 \multicolumn{1}{c|}{} & \multicolumn{1}{c}{GSM-IC2-1K} & \multicolumn{1}{c}{GSM-ICM-1K} \\ 
 \hline
 Zero-Shot-CoT & $87.0$ & $82.0$ \\ \hline
 Zero-Shot-CoT + Refine & $89.2$ & $84.8$ \\
 Zero-Shot-CoT + I$^3$C & $\mathbf{92.7}$ & $\mathbf{88.6}$ \\
\toprule
\end{tabular}
\end{adjustbox}
\label{tab:refine}
\end{table}

\subsection{Ablation Studies}

\paragraph{How does demonstration construction affect I$^3$C-Select?}
In I$^3$C-Select, we select the $K$ most confusing problems and their reasoning paths as demonstrations and named this demonstration construction method ``High''.
To verify the effectiveness of the demonstration construction method, we also consider: (1) ``Low'', where we select the $K$ problems with the lowest confusion scores and their reasoning paths as demonstrations, and (2) ``Medium'', where we randomly select $K$ problems and their reasoning paths as demonstrations. 
We set $K$ to 8 throughout our experiments.
As shown in Figure~\ref{fig:easy-medium}, selecting more confusing problems and their reasoning paths as demonstrations can effectively improve the model's performance.

\begin{table}[t]
\renewcommand\arraystretch{1.15}
\caption{Accuracy (\%) comparison of different demonstration construction methods.}
\centering
\begin{adjustbox}{width=0.9\columnwidth,center}
\begin{tabular}{l|p{1.6cm}<{\centering}p{1.6cm}<{\centering}}
\bottomrule
 \multicolumn{1}{l|}{\multirow{2}{*}{\makecell{Method\\(GPT-3.5-Turbo)}}} & \multicolumn{2}{c}{Dataset} 
 \\ \cline{2-3}
 \multicolumn{1}{c|}{} & \multicolumn{1}{c}{GSM-IC2-1K} & \multicolumn{1}{c}{GSM-ICM-1K} \\ 
 \hline
 Complex-CoT & $84.3$ & $83.0$ \\ \hline
 I$^3$C-Select - I$^3$C & $\mathbf{92.7}$ & $\mathbf{89.5}$ \\
\toprule
\end{tabular}
\end{adjustbox}
\label{tab:demonstration}
\end{table}

\paragraph{Instructing to ignore irrelevant conditions vs. refining problems to eliminate irrelevant conditions.}
In Zero-Shot-CoT+I$^3$C, we use I$^3$C instruction to instruct LLMs to identify and ignore irrelevant conditions in the MWP solving process.
In addition, we can refine the given problem to eliminate irrelevant conditions based on the verification outputs generated in \cref{sec:i3c_instruction}, and solve the refined problem using the Zero-Shot-CoT method (i.e., Zero-Shot-CoT+Refine).
As shown in Table \ref{tab:refine}, 
Zero-Shot-CoT+Refine ($89.2$ and $84.8$) substantially outperforms Zero-Shot-CoT ($87.0$ and $82.0$) on GSM-IC2-1K and GSM-ICM-1K, respectively.
This highlights that the generated verification outputs can explicitly identify irrelevant conditions in the problem description.
Furthermore, Zero-Shot-CoT+I$^3$C consistently outperforms Zero-Shot-CoT+Refine.
This is mainly because the identified irrelevant conditions may contain some useful conditions. When we refine the given problem, we may eliminate some useful conditions, resulting in an incorrect answer.
Instructing the LLM to ignore irrelevant conditions can effectively alleviate the problem of losing useful conditions during problem refinement.
Case studies are provided in Appendix \ref{sec:appendix_results}.

\paragraph{Comparison of different demonstration construction methods.}
To evaluate the effectiveness of the demonstration construction methods, we also consider I$^3$C-Select - I$^3$C, which selects the 8 most confusing problems and their reasoning paths as demonstrations, without including the I$^3$C instruction in the prompt.
Table~\ref{tab:demonstration} shows that I$^3$C-Select - I$^3$C ($92.7$ and $89.5$) significantly outperforms Complex-CoT ($84.3$ and $83.0$) on GSM-IC2-1K and GSM-ICM-1K, respectively.
These results suggest that selecting the most confusing problems and their reasoning paths as demonstrations is a more effective demonstration construction method.

\section{Conclusion}
\label{sec:conclusion}
In this study, we introduce a plug-and-play module, I$^3$C, which can be added to any CoT prompting methods to enhance LLMs' ability to explicitly identify and ignore irrelevant conditions in the mathematical problem-solving process.
Moreover, we propose a novel few-shot prompting method, I$^3$C-Select, which selects the most confusing problems and their corresponding reasoning paths as demonstrations.
Extensive experiments on eight math word problem datasets demonstrate the effectiveness and efficiency of our proposed method.

\section*{Acknowledgments}
Zhenyu Wu was a visiting student at the University of Notre Dame, advised by Meng Jiang. His visit was financially supported by Xi'an Jiaotong University. Chao Shen is the corresponding author.

\bibliography{reference}

\appendix
\section{Appendix}
\subsection{Datasets}
\label{sec:appendix_datasets}
We use eight math word problem datasets for assessing prompting method quality. The statistics of the datasets are shown in Table~\ref{tab:data}. 
All of these datasets are accessible under the MIT License. 
We give a brief description of the datasets used below:
\begin{list}{$-$}{\leftmargin=2em \itemindent=0em}
  \item SingleEq \citep{koncel-2015-singleq} contains a set of grade-school algebra word problems. Each problem may involve multiple math operations including multiplication, division, subtraction, and addition.
  \item AddSub \citep{Hosseini-2014-AddSub} consists of math word problems on addition and subtraction for third, fourth, and fifth graders.
  \item SVAMP \citep{patel-2021-svamp} consists of one-unknown math word problems which can be solved by expressions requiring no more than two operators.
  \item GSM8K \citep{karl-2021-gsm8k} consists of high quality grade school math word problems created by human problem writers. These problems take between $2$ and $8$ steps to solve, and solutions primarily involve performing a sequence of elementary calculations using basic arithmetic operations to reach the final answer.
  \item AQuA \citep{ling-etal-2017-aqua} consist of multiple option math questions covering a broad range of topics and difficulty levels.
  \item MATH \citep{dan-etal-2021-MATH} is a challenging datasets consisting of 12k problems within 7 categories testing the models' advanced math and science reasoning. The problems in this dataset are very hard as they come from mathematics competitions written in \LaTeX.
  \item GSM-IC \citep{freda-2023-gsm8k-ic} is an arithmetic reasoning dataset with irrelevant conditions in the problem description. It is divided into two splits: GSM-IC2, consisting of problems requiring two steps to solve, and GSM-ICM, consisting of problems requiring more than two steps to solve. Being mindful of the experiment costs, we uniformly sample $1,000$ examples from the GSM-IC2 dataset (denoted by GSM-IC2-1K) and $1,000$ examples from the GSM-ICM dataset (denoted by GSM-ICM-1K) for evaluation and analysis purposes throughout this paper.
\end{list}

\begin{table}[t]
\centering
\caption{Dataset description. The last column indicates the percentage of problems with irrelevant conditions in the problem description.}
\renewcommand\arraystretch{1.1}
  \resizebox{1.0\columnwidth}{!}{
  \begin{tabular}{l|ccc}
    \bottomrule
    Dataset & \# Problems & Avg.\# Words & Irrelevant Condition \\
    \hline
    SingleEq & $508$ & $27.4$ & $\ \ \ 0.0\%$ \\
    AddSub & $395$ & $31.5$ & \ $30.9\%$ \\
    SVAMP & $1,000$ & $31.8$ & \ $36.7\%$ \\
    GSM8K & $1,319$ & $46.9$ & \ \ \  $6.2\%$ \\
    AQuA & $254$ & $51.9$ & $  \ 14.2\%$ \\
    MATH & $500$ & $68.6$ & $  \ \  \ 3.8\%$ \\
    GSM-IC2-1K & $1,000$ & $41.8$ & $ 100.0\%$ \\
    GSM-ICM-1K & $1,000$ & $61.4$ & $ 100.0\%$ \\
    \toprule
\end{tabular} }
\label{tab:data}
\end{table}

\subsection{Baselines}
\label{sec:appendix_baselines}
As we study how to prompt large language models to solve math word problems, we employ seven prompting baselines. We give a brief description of the baselines used below:
\begin{list}{$-$}{\leftmargin=2em \itemindent=0em}
  \item Direct \citep{Kojima-2022-CoT} is a baseline that utilizes the symbolic reasoning ability of large language models. By simply adding the sentence ``\emph{The answer is}'' after the problem of interest, which instructs the large language model to generate the answer to the problem.
  \item Zero-Shot-CoT \citep{Kojima-2022-CoT} is a Chain-of-Thought prompting method. By adding ``\emph{Let's think step by step}'' to the problem to elicit the large language model to generate reasoning path leading to the final answer.
  \item Plan-and-Solve (PS) \citep{wang-2023-ps} replaces the sentence ``\emph{Let's think step by step}'' with ``\emph{Let's first understand the problem and devise a plan to solve the problem. Then let's carry out the plan and solve the problem step by step}'' to address the missing step issue in Zero-Shot-CoT.
  \item Instruct-CoT \citep{freda-2023-gsm8k-ic} adds the sentence ``\emph{Feel free to ignore irrelevant conditions in the problem description.}'' before the problem of interest, which instructs the large language model to ignore irrelevant information in the problem description.
  \item  Manual-CoT \citep{wei-2022-cot} is a few-shot prompting method. By representing manual designed demonstrations that solve the corresponding problems with intermediate reasoning steps in the prompts, Manual-CoT elicits multi-step reasoning ability of LLMs.
  \item  Auto-CoT \citep{zhang-2023-autocot} automatically constructs demonstrations with questions and reasoning paths from the test set to eliminate manual designs in Manual-CoT.
  \item  Complex-CoT \citep{fu-2023-complexity} is a few-shot prompting method that selects the most complex problems and their generated reasoning paths as demonstrations.
  \item  PAL \citep{Gao-2023-pal} is a few-shot prompting method that generates programming language statements and uses a program interpreter to execute the generated program to get final answers.
\end{list}

\begin{table}[htbp]
\centering
\caption{All prompts used in experiments. $Q$ represents the problem to be solved. $I$ represents the I$^3$C instruction that instructs LLMs to identify and ignore irrelevant conditions in the problem description. The demonstrations of Manual-CoT is from its original paper \citep{wei-2022-cot}.}
\renewcommand\arraystretch{1.1}
  \resizebox{\columnwidth}{!}{
  \begin{tabular}{ll}
    \bottomrule
    Method & Prompt \\
    \hline
    Direct & \makecell[l]{\emph{Q: $Q$} \\  \emph{A: The answer is.}} \\
    \hline
    Direct + I$^3$C & \makecell[l]{$I$ \\ \emph{Q: $Q$} \\  \emph{A: The answer is.}} \\
    \hline
    Zero-Shot-CoT & \makecell[l]{\emph{Q: $Q$} \\  \emph{A: Let's think step by step.}} \\
    \hline
    Zero-Shot-CoT + I$^3$C & \makecell[l]{$I$ \\ \emph{Q: $Q$} \\  \emph{A: Let's think step by step.}} \\
    \hline
    PS & \makecell[l]{\emph{Q: $Q$} \\  \emph{A: Let's first understand the problem} \\ \emph{and devise a plan to solve the problem.} \\ \emph{Then, let's carry out the plan and solve} \\ \emph{the problem step by step.}} \\
    \hline
    PS + I$^3$C & \makecell[l]{$I$ \\ \emph{Q: $Q$} \\  \emph{A: Let's first understand the problem} \\ \emph{and devise a plan to solve the problem.} \\ \emph{Then, let's carry out the plan and solve} \\ \emph{the problem step by step.}} \\
    \hline
    Instruct-CoT & \makecell[l]{\emph{Feel free to ignore irrelevant conditions}  \\ \emph{in the problem description.} \\ \emph{Q: $Q$} \\  \emph{A: Let's think step by step.}} \\
    \hline
    Instruct-CoT + I$^3$C & \makecell[l]{$I$ \\ \emph{Feel free to ignore irrelevant conditions}  \\ \emph{in the problem description.} \\ \emph{Q: $Q$} \\  \emph{A: Let's think step by step.}} \\
    \hline
    Manual-CoT & \makecell[l]{\{hand-crafted demonstrations\} \\ \emph{Q: $Q$} \\  \emph{A:}} \\
    \hline
    Manual-CoT + I$^3$C & \makecell[l]{$I$ \\ \{hand-crafted demonstrations\} \\\emph{Q: $Q$} \\  \emph{A:}} \\
    \hline
    Auto-CoT & \makecell[l]{\{demonstrations\} \\ \emph{Q: $Q$} \\  \emph{A:}} \\
    \hline
    Auto-CoT + I$^3$C & \makecell[l]{$I$ \\ \{demonstrations\} \\ \emph{Q: $Q$} \\  \emph{A:}} \\
    \hline
    Complex-CoT & \makecell[l]{\{demonstrations\} \\ \emph{Q: $Q$} \\  \emph{A:}} \\
    \hline
    Complex-CoT + I$^3$C & \makecell[l]{$I$ \\ \{demonstrations\} \\ \emph{Q: $Q$} \\  \emph{A:}} \\
    \hline
    PAL & \makecell[l]{\{demonstrations\} \\ \emph{Q: $Q$} \\  \emph{A:}} \\
    \hline
    I$^3$C-Select (Ours) & \makecell[l]{$I$ \\ \{demonstrations\} \\ \emph{Q: $Q$} \\  \emph{A:}} \\
    \toprule
\end{tabular} }
\label{tab:prompts}
\end{table}

\subsection{Metrics}
\label{sec:appendix_metrics}
We use accuracy to evaluate the performance of different prompting methods. Since large language models cannot perform the computation precisely (especially with high-precision floats), we consider an answer to be correct if and only if the absolute error between the answer and the gold answer is less than $1 \times 10^{-5}$. Let $\mathcal{P}$ be a set of problems, the accuracy of the prompting method is
\begin{scriptsize}
\begin{align}
\text { Accuracy }&=\frac{1}{|\mathcal{P}|} \sum_{Q \in \mathcal{P}} \mathds{1}\left(a^{(\text {final })}, a^{(\text {gold })}\right)\nonumber \\
\mathds{1}\left(a^{(\text {final })}, a^{(\text {gold })}\right)&= \begin{cases}1, & \text{if} \  \operatorname{Abs}\left(a^{(\text {final })}-a^{(\text {gold })}\right) < 1 \times 10^{-5} \\
0, & \text{if} \ \operatorname{Abs}\left(a^{(\text {final })}-a^{(\text {gold })}\right) \geq 1 \times 10^{-5}\end{cases}\nonumber
\label{eqacc}
\end{align}
\end{scriptsize}
where $a^{(\text {gold })}$ is the gold answer to question $Q$, $a^{(\text {final })}$ is the model-generated answer to question $Q$, and $\operatorname{Abs}(\cdot)$ is the absolute value function.

\subsection{Full prompts in experiments}
\label{sec:appendix_prompts}
We list the prompts for all experiments in Table~\ref{tab:prompts}.

\begin{table*}[t]
\renewcommand\arraystretch{1.15}
\caption{Accuracy (\%) comparison on six MWP datasets. I$^3$C indicates that instructs LLMs to identify and ignore irrelevant conditions. Adding the I$^3$C instruction to CoT prompting methods effectively improves performance. Selecting the most confusing problems and their generated reasoning paths as demonstrations for few-shot learning (i.e., I$^3$C-Select) achieves state-of-the-art performance on all six MWP datasets.}
\centering
\begin{adjustbox}{width=\textwidth,center}
\begin{tabular}{l|p{1.6cm}p{1.6cm}p{1.6cm}p{1.6cm}p{1.6cm}p{1.6cm}}
\bottomrule
 \multicolumn{1}{l|}{\multirow{2}{*}{\makecell[c]{Method \\ (UL2-20B)}}} & \multicolumn{6}{c}{Dataset} 
 \\ \cline{2-7}
 \multicolumn{1}{c|}{} & \multicolumn{1}{l}{AddSub} & \multicolumn{1}{l}{SVAMP} & \multicolumn{1}{l}{GSM8K} & \multicolumn{1}{l}{SingleEq} & \multicolumn{1}{l}{GSM-IC2-1K} & \multicolumn{1}{l}{GSM-ICM-1K} \\ 
 \hline
 Direct & $28.6$ & $16.9$ & $5.0$ & $21.7$ & $12.9$ & $9.5$ \\
 Direct + I$^3$C & $33.9 (+5.3)$ & $27.8 (+10.9)$ & $9.8 (+4.8)$ & $32.7 (+11.0)$ & $21.3 (+8.4)$ & $13.2 (+3.7)$ \\
 \hline
 
 Zero-Shot-CoT & $32.9$ & $29.5$ & $22.7$ & $38.8$ & $29.6$ & $25.5$ \\
 Zero-Shot-CoT + I$^3$C & $36.7 (+3.8)$ & $30.5 (+1.0)$ & $22.7 (+0.0)$ & $40.0 (+1.2)$ & $40.6 (+11.0)$ & $27.6 (+2.1)$ \\
 \hline
 
 PS & $30.0$ & $26.7$ & $21.2$ & $36.6$ & $27.4$ & $24.9$ \\
 PS + I$^3$C & $31.9 (+1.9)$ & $28.4 (+1.7)$ & $21.3 (+0.1)$ & $40.0 (+3.4)$ & $32.4 (+5.0)$ & $26.0 (+1.1)$ \\
 \hline
 
 Instruct-CoT & $34.7$ & $31.2$ & $23.5$ & $40.0$ & $33.8$ & $26.4$ \\
 Instruct-CoT + I$^3$C & $35.4 (+0.7)$ & $31.5 (+0.3)$ & $21.2 (-2.3)$ & $41.1 (+1.1)$ & $40.0 (+6.2)$ & $28.6 (+2.2)$ \\
 
 \hline
 Manual-CoT & $34.9$ & $31.7$ & $25.2$ & $43.3$ & $35.4$ & $28.0$ \\
 Manual-CoT + I$^3$C & $39.0 (+4.1)$ & $28.1 (-3.6)$ & $22.2 (-3.0)$ & $42.9 (-0.4)$ & $43.0 (+7.6)$ & $28.5 (+0.5)$ \\
 \hline
 
 Auto-CoT & $36.7$ & $31.9$ & $24.5$ & $41.9$ & $35.0$ & $29.4$ \\
 Auto-CoT + I$^3$C & $39.5 (+2.8)$ & $28.7 (-3.2)$ & $24.7 (+0.2)$ & $43.6 (+1.7)$ & $41.1 (+6.1)$ & $30.1 (+0.7)$ \\
 \hline
 
 
 I$^3$C-Select (Ours) & $\mathbf{39.7}$ & $\mathbf{34.6}$ & $\mathbf{27.5}$ & $\mathbf{44.1}$ & $\mathbf{46.0}$ & $\mathbf{35.9}$ \\
\toprule
\end{tabular}
\end{adjustbox}
\label{tab:mainresult-lm}
\end{table*}

\begin{figure*}[t]
  \centering
  \includegraphics[width=0.9\textwidth]{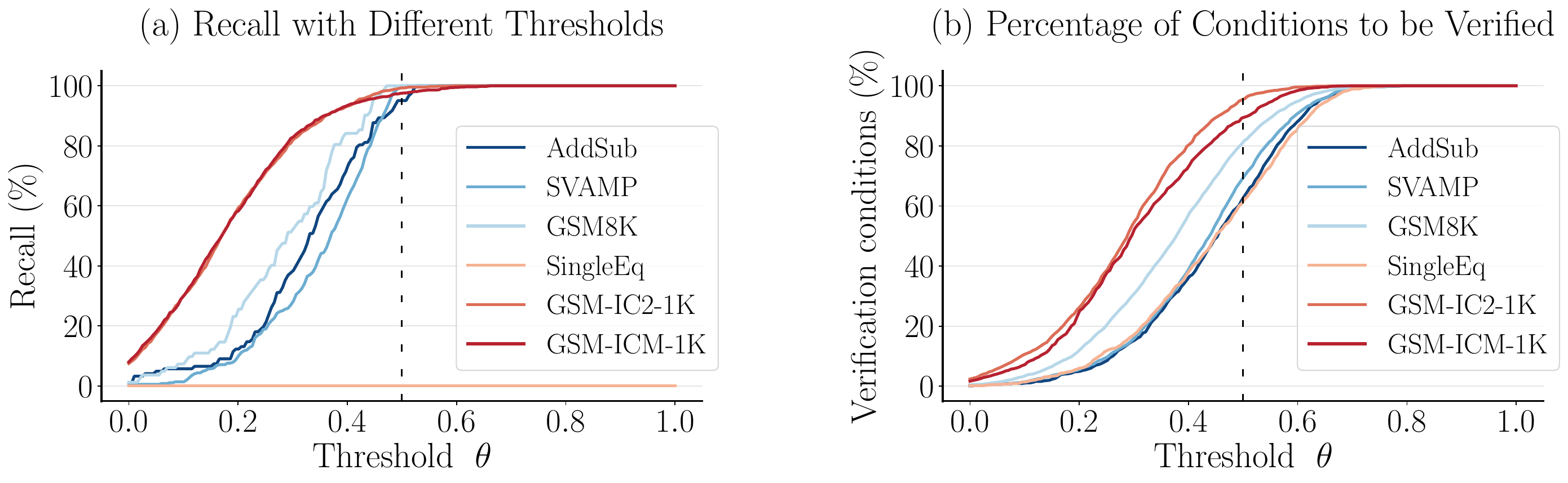}
  \caption{Hyperparameter analysis. (a) As the threshold increases, the recall scores of identified irrelevant condition candidates first increase and then remain unchanged for all datasets except SingleEq. (b) As the threshold increases, the percentage of conditions to be verified first increases and then remains unchanged for all datasets.}
  \label{fig:threshold}
\end{figure*}

\subsection{Additional Experimental Results}
\label{sec:appendix_results}
\paragraph{Does I$^3$C instruction work with weaker LMs?} 
In all our experiments in \cref{sec:experiments}, we use GPT-3 (text-davinci-003) and GPT-3.5-Turbo as backend LLMs, but can I$^3$C instruction work with weaker LMs?
We compare CoT prompting methods with adding the I$^3$C instruction to CoT prompting methods when use UL2-20B as backend LM.
Note that UL2-20B is a weaker LMs with 20 billion parameters, but GPT3 has 175 billion parameters.
As shown in Table~\ref{tab:mainresult-lm}, even though the absolute accuracies of UL2-20B are lower, adding the I$^3$C instruction to CoT prompting methods effectively improves MWP solving performance, and I$^3$C-Select achieves consistent performance improvements on MWP datasets.
This shows that I$^3$C instruction can work with weaker LMs. 

\begin{table*}[t]
\caption{Case study of answers generated by different methods. Irrelevant conditions in the problem description is highlighted with red color, and the LLM-generated final answer is highlighted with yellow color.}
\resizebox{0.98\textwidth}{!}{
\begin{tabularx}{\textwidth}{p{2.5cm}|X}
\toprule
   \textbf{Problem} & Granger went to the grocery store. He saw that the Spam is \$3 per can, the peanut butter is \$5 per jar, and the bread is \$2 per loaf. \sethlcolor{pink}\hl{Jack owns 5 bottles of water with each worth \$2.} If Granger bought 12 cans of spam, 3 jars of peanut butter, and 4 loaves of bread, how much is the total amount he paid? \\
\midrule
   \textbf{Gold Answer} & $3 \times 12 + 5 \times 3 + 2 \times 4 = \ $\textbf{59} \\
\midrule
   \textbf{Auto-CoT} & Granger bought 12 cans of spam. 12 * \$3 = \$36. Granger bought 3 jars of peanut butter. 3 * \$5 = \$15. Granger bought 4 loaves of bread. 4 * \$2 = \$8. Jack owns 5 bottles of water. 5 * \$2 = \$10. The total amount he paid is \$36 + \$15 + \$8 + \$10 = \$69. The answer is \sethlcolor{yellow}\hl{69}. \textcolor{red}{\XSolidBrush}\\\midrule
   \textbf{Instruct-CoT} & Granger bought 12 cans of Spam, so that’s 12 x \$3 = \$36. He bought 3 jars of peanut butter, so that’s 3 x \$5 = \$15. He bought 4 loaves of bread, so that’s 4 x \$2 = \$8. Jack owned 5 bottles of water, so that’s 5 x \$2 = \$10. Adding them all up, the total amount Granger paid is \$36 + \$15 + \$8 + \$10 = \$\sethlcolor{yellow}\hl{69}. \textcolor{red}{\XSolidBrush}\\\midrule 

   \textbf{Complex-CoT} & The cost of 12 cans of spam is 12 x \$3 = \$36. The cost of 3 jars of peanut butter is 3 x \$5 = \$15. The cost of 4 loaves of bread is 4 x \$2 = \$8. The cost of 5 bottles of water is 5 x \$2 = \$10. Therefore, the total amount Granger paid is \$36 + \$15 + \$8 + \$10 = \$69. The answer is \$\sethlcolor{yellow}\hl{69}. \textcolor{red}{\XSolidBrush}\\\midrule

   \textbf{I$^3$C-Select} & First, calculate the cost for each item that Granger bought: 12 cans of Spam = 12 x \$3 = \$36; 3 jars of peanut butter = 3 x \$5 = \$15; 4 loaves of bread = 4 x \$2 = \$8. Then, calculate the total cost of all the items: \$36 + \$15 + \$8 = \$\sethlcolor{yellow}\hl{59}. \textcolor{red}{\Checkmark} \\ 
\bottomrule
\end{tabularx} }
\vspace{-0.05in}
\label{tab:casestudy}
\end{table*}

\paragraph{Hyperparameter setup for I$^3$C instruction.} 
To compare the identification accuracy of irrelevant conditions at different thresholds $\theta$, we employ the recall metric. 
Recall is computed by dividing the number of correctly identified irrelevant conditions by the total number of irrelevant conditions. 
As shown in Figure~\ref{fig:threshold}a, as the threshold increases, the recall score for the irrelevant condition identification first increases and then remains unchanged in all datasets except SingleEq.
The exception is due to the problems in SingleEq not containing irrelevant conditions, the recall score remains $0$ regardless of the threshold.
Notably, when the threshold is $0.5$, we can accurately identify $98.6\%$ of irrelevant conditions.
Furthermore, as shown in Figure~\ref{fig:threshold}b, the percentage of conditions to be verified initially increases and then stabilizes across all datasets as the threshold increases. 
The percentage of conditions to be verified is calculated by dividing the number of conditions in the set of irrelevant condition candidates by the total number of conditions. 
When the threshold is $0.5$, the LLM is required to verify $66.6\%$ of conditions on average, and the average number of verification calls per MWP made to the LLM is approximately $2.18$.
Overall, we set the threshold $\theta$ to $0.5$ to achieve a trade-off between efficiency and effectiveness.

\paragraph{I$^3$C instruction enhances robustness against irrelevant conditions.} 
Table~\ref{tab:casestudy} shows that apart from I$^3$C-Select, the other methods fail to provide accurate answers to the given problem.
Instruct-CoT, Complex-CoT, and Auto-CoT produce incorrect answers due to the incorporation of irrelevant conditions in the MWP solving process.
In contrast, I$^3$C-Select explicitly identifies and ignores irrelevant conditions in the MWP solving process.

\begin{table*}[t]
\caption{Example outputs by Zero-Shot-CoT + Refine and Zero-Shot-CoT + I$^3$C for GSM-ICM-1K. Irrelevant conditions in the problem description is highlighted with red color.}
\renewcommand\arraystretch{1.35}
\resizebox{0.98\textwidth}{!}{
\begin{tabularx}{\textwidth}{p{2.6cm}|p{6.4cm}|p{6.0cm}}
\hline   

\textbf{Problem} & \multicolumn{2}{p{12.8cm}}{A cobbler can mend 3 pairs of shoes in an hour. From Monday to Thursday, the cobbler works for 8 hours each day, and on Friday, he only works from 8am to 11am. \sethlcolor{pink}\hl{Last week, Tom repaired 10 more umbrellas than the cobbler.} How many pairs of shoes can the cobbler mend in a week?}  \\ \hline

\textbf{Gold Answer} & \multicolumn{2}{X}{$(4 \times 8 + 11 - 8) \times 3 = 105$} \\ \hline

\textbf{Verification Outputs} & \multicolumn{2}{p{1cm}}{\begin{tabular}[c]{@{}p{12.8cm}@{}}
$\blacksquare$ Yes, the condition A cobbler can mend 3 pairs of shoes in an hour is \textbf{relevant} to the calculation process of the problem How many pairs of shoes can the cobbler mend in a week? Without the condition, it would be impossible to calculate the number of pairs of shoes the cobbler can mend in a week.  \\
$\blacksquare$ The condition on Friday, he only works from 8am to 11am is \textbf{not relevant} to the calculation process. \\
$\blacksquare$ The condition Last week, Tom repaired 10 more umbrellas than the cobbler. is \textbf{not relevant} to the calculation process of the problem.
\end{tabular}} \\ \hline

\textbf{Refined Problem} & \multicolumn{2}{p{12.8cm}}{A cobbler can mend 3 pairs of shoes in an hour. From Monday to Thursday, the cobbler works for 8 hours each day. How many pairs of shoes can the cobbler mend in a week?}  \\ \hline

\textbf{Reasoning Path} & \multicolumn{1}{c|}{\textbf{Zero-Shot-CoT + Refine}} & \multicolumn{1}{c}{\textbf{Zero-Shot-CoT + I$^3$C}} \\ 
\cline{2-3} 
 
& First, we need to calculate how many pairs of shoes the cobbler can mend in 4 days (Monday to Thursday). Since he works 8 hours a day, and can mend 3 pairs of shoes per hour, he can mend 8 x 3 = 24 pairs of shoes in 4 days. This gives us 24 pairs of shoes in a week.
& Step 1: Calculate the number of hours the cobbler works in a week. From Monday to Thursday, the cobbler works for 8 hours each day. That's 8 x 4 = 32 hours. On Friday, he only works from 8am to 11am. That's 3 hours. So in a week, the cobbler works 32 + 3 = 35 hours. Step 2: Calculate the number of pairs of shoes the cobbler can mend in a week. The cobbler can mend 3 pairs of shoes in an hour. So in 35 hours, the cobbler can mend 35 x 3 = 105 pairs of shoes. Step 3: Determine if the condition is relevant to the calculation process. The fact that Tom repaired 10 more umbrellas than the cobbler does not affect the calculation process. Therefore, the cobbler can mend 105 pairs of shoes in a week.   
\\ \hline

\textbf{Final Answer} & 24 \textcolor{red}{\XSolidBrush} & 105 \textcolor{red}{\Checkmark} \\   

\hline                                
\end{tabularx} }
\label{tab:case_refine}
\vspace{-0.3in}
\end{table*}

\paragraph{Instructing to ignore irrelevant conditions vs. refining problems to eliminate irrelevant conditions.}
In Zero-Shot-CoT+I$^3$C, we use I$^3$C instruction to instruct LLMs to identify and ignore irrelevant conditions in the MWP solving process.
In addition, we can refine the given problem to eliminate irrelevant conditions based on the verification outputs generated in \cref{sec:i3c_instruction}, and solve the refined problem using the Zero-Shot-CoT method (i.e., Zero-Shot-CoT+Refine).
For example, as shown in Table \ref{tab:case_refine}, the condition ``\emph{On Friday, he only works from 8am to 11am.}'' and the condition ``\emph{Last week, Tom repaired 10 more umbrellas than the cobbler.}'' are identified as the irrelevant conditions.
By eliminating these identified irrelevant conditions, we get the refined problem ``\emph{A cobbler can mend 3 pairs of shoes in an hour. From Monday to Thursday, the cobbler works for 8 hours each day. How many pairs of shoes can the cobbler mend in a week?}''.
Obviously, in this case, we incorrectly identified the condition ``\emph{On Friday, he only works from 8am to 11am.}'' as an irrelevant condition.
Eliminating this condition would result in losing useful conditions in the problem refinement process, resulting in an incorrect answer.
In contrast, instructing the LLM to ignore irrelevant conditions can effectively alleviate the problem of losing useful conditions during problem refinement, and can effectively enhance the MWP solving performance.



\subsection{Sample Predictions on MWP Datasets}
\label{sec:appendix_case}
We present case studies in \Cref{tab:case_1,tab:case_2,tab:case_3,tab:case_4,tab:case_5,tab:case_6}.
It is evident that I$^3$C instructs LLMs to explicitly identify and ignore irrelevant conditions in the problem statement.
Adding I$^3$C instruction to the CoT prompting methods effectively helps LLMs avoid confusion and improve reasoning paths.

\begin{table*}[t]
\centering
\caption{Example outputs by Zero-Shot-CoT, Zero-Shot-CoT + I$^3$C and I$^3$C-Select. Irrelevant conditions in the problem description is highlighted with red color, and the LLM-generated final answer is highlighted with yellow color.}
\renewcommand\arraystretch{1.35}
\resizebox{\textwidth}{!}{
\begin{tabularx}{\textwidth}{p{2.5cm}|X}
\hline   

\textbf{Problem} & Megan is an actress. She was the lead actress in 80\% of her work. In total, Megan participated in 100 plays. \sethlcolor{pink}\hl{Tom bought the tickets for 300 plays.} How many times Megan was not the lead actress? \\ \hline

\textbf{Gold Answer} & $100 \times (1-80\%) = 20$ \\ \hline

\textbf{Verification Outputs} & \begin{tabular}[c]{@{}p{12.8cm}@{}}$\blacksquare$ Megan participated in 100 plays. That means she was the lead actress in 80 of them. That leaves 20 plays where she was not the lead actress. Tom bought tickets to 300 plays. That means Megan was the lead actress in 240 of them, and she was not the lead actress in 60 of them. So, the answer is 60 times. \textcolor{teal!70}{The condition She was the lead actress in 80\% of her work is relevant to the calculation process, as it tells us that 80 out of 100 plays she was the lead actress.}\\ $\blacksquare$ Megan was the lead actress in 80\% of her plays, so she was the lead actress in 80 plays. Since she participated in a total of 100 plays, that means she was not the lead actress in the remaining 20 plays. \textcolor{teal!70}{The condition that Tom bought tickets for 300 plays is irrelevant to the calculation process.}\end{tabular} \\ \hline

\multirow{6}{*}{\textbf{Reasoning Path}} 
& \textbf{Zero-Shot-CoT} \\
& Megan was the lead actress in 80 plays.  100 - 80 = 20   Tom bought the tickets for 300 plays.  Megan was not the lead actress in the remaining \sethlcolor{yellow}\hl{200} plays. \textcolor{red}{\XSolidBrush} \\ \cline{2-2} 
& \textbf{Zero-Shot-CoT + I$^3$C} \\ 
& Megan was the lead actress in 80\% of her work, which means she was the lead actress in 80 plays. That leaves 20 plays in which she was not the lead actress. Tom bought tickets for 300 plays, but that information does not affect the calculation of how many times Megan was not the lead actress. Therefore, the answer is \sethlcolor{yellow}\hl{20} times.  \textcolor{red}{\Checkmark} \\ \cline{2-2} 
& \textbf{I$^3$C-Select} \\ 
& Megan was the lead actress in 80\% of her work, so she was the lead actress in 80 plays. That leaves 20 plays in which she was not the lead actress. Tom bought the tickets for 300 plays, but that information does not affect the calculation of how many times Megan was not the lead actress. Therefore, the answer is \sethlcolor{yellow}\hl{20} times. \textcolor{red}{\Checkmark}  \\

\hline                                
\end{tabularx} }
\label{tab:case_1}
\end{table*}

\begin{table*}[t]
\centering
\caption{Example outputs by Direct, Direct + I$^3$C and I$^3$C-Select. Irrelevant conditions in the problem description is highlighted with red color, and the LLM-generated final answer is highlighted with yellow color.}
\renewcommand\arraystretch{1.35}
\resizebox{\textwidth}{!}{
\begin{tabularx}{\textwidth}{p{2.5cm}|X}
\hline   

\textbf{Problem} & Sally eats 2 sandwiches on Saturday and 1 sandwich on Sunday. \sethlcolor{pink}\hl{Jack bought 2 more pieces of bread than Sally yesterday.} If each sandwich uses 2 pieces of bread, how many pieces of bread does Sally eat across Saturday and Sunday? \\ \hline

\textbf{Gold Answer} & $(2 + 1) \times 2 = 6$ \\ \hline

\textbf{Verification Outputs} & \begin{tabular}[c]{@{}p{12.8cm}@{}}$\blacksquare$ Sally eats 2 sandwiches on Saturday, so she has eaten 4 pieces of bread.  On Sunday, she eats 1 sandwich, so she has eaten 2 more pieces of bread.  Jack bought 2 more pieces of bread than Sally, so Sally has eaten 6 pieces of bread across Saturday and Sunday.   Therefore, the answer is 6 pieces of bread.   \textcolor{teal!70}{The condition of 1 sandwich on Sunday is relevant to the calculation process, since it tells us how many pieces of bread Sally has eaten on Sunday.}  \\
$\blacksquare$ Sally ate 2 sandwiches on Saturday, so she used 4 pieces of bread on Saturday.  On Sunday, Sally ate 1 sandwich, so she used 2 pieces of bread on Sunday.  Therefore, Sally ate 6 pieces of bread across Saturday and Sunday.  \textcolor{teal!70}{The condition Jack bought 2 more pieces of bread than Sally is not relevant to the calculation process, as it does not affect the number of pieces of bread Sally ate across Saturday and Sunday.}
\end{tabular} \\ \hline

\multirow{6}{*}{\textbf{Reasoning Path}} 
& \textbf{Direct} \\
& Sally eats a total of \sethlcolor{yellow}\hl{4} pieces of bread across Saturday and Sunday. \textcolor{red}{\XSolidBrush} \\ \cline{2-2} 
& \textbf{Direct + I$^3$C} \\ 
& 6 pieces of bread. Sally ate 2 sandwiches on Saturday, which used 4 pieces of bread, and 1 sandwich on Sunday, which used 2 pieces of bread, so she ate \sethlcolor{yellow}\hl{6} pieces of bread across Saturday and Sunday. The condition that Jack bought 2 more pieces of bread than Sally is not relevant to the calculation process. \textcolor{red}{\Checkmark} \\ \cline{2-2} 
& \textbf{I$^3$C-Select} \\ 
& Sally ate 2 sandwiches on Saturday, so she used 4 pieces of bread on Saturday. On Sunday, Sally ate 1 sandwich, so she used 2 pieces of bread on Sunday. Therefore, Sally ate \sethlcolor{yellow}\hl{6} pieces of bread across Saturday and Sunday. The condition Jack bought 2 more pieces of bread than Sally is not relevant to the calculation process, as it does not affect the number of pieces of bread Sally ate across Saturday and Sunday. \textcolor{red}{\Checkmark}  \\

\hline                                
\end{tabularx} }
\label{tab:case_2}
\end{table*}

\begin{table*}[t]
\centering
\caption{Example outputs by PS, PS + I$^3$C and I$^3$C-Select. Irrelevant conditions in the problem description is highlighted with red color, and the LLM-generated final answer is highlighted with yellow color.}
\renewcommand\arraystretch{1.35}
\resizebox{\textwidth}{!}{
\begin{tabularx}{\textwidth}{p{2.5cm}|X}
\hline   

\textbf{Problem} & Holden's current master bedroom is 309 sq ft and his master bath is 150 sq ft. \sethlcolor{pink}\hl{The basketball field of Tom is 200 sq ft larger than Holden's bathroom.} If Holden wants to add a home office/personal gym divided space that is twice as large as his bedroom and bathroom, how much sq ft will this new room have? \\ \hline

\textbf{Gold Answer} & $(309 + 105) \times 2 = 918$ \\ \hline

\textbf{Verification Outputs} & \begin{tabular}[c]{@{}p{12.8cm}@{}}
$\blacksquare$ The master bedroom is 309 sq ft and the master bathroom is 150 sq ft.  Therefore, the combined area of the bedroom and bathroom is 459 sq ft.  If this space is to be doubled for the new home office/personal gym, then the new room will have 918 sq ft.  \textcolor{teal!70}{The condition Holden's current master bedroom is 309 sq ft is relevant to the problem, as it is used to determine the combined area of the bedroom and bathroom (459 sq ft).} This combined area is then doubled to get the total area of the new room (918 sq ft).  \\
$\blacksquare$ Holden's bedroom is 309 sq ft.  Holden's bathroom is 150 sq ft.  The total sq ft of Holden's bedroom and bathroom is 459 sq ft.  We need to find a space that is twice as large as Holden's bedroom and bathroom combined.  Therefore, the new room will be 2 x 459 sq ft = 918 sq ft.   \textcolor{teal!70}{The basketball field of Tom is 200 sq ft larger than Holden's bathroom so it is not relevant to the calculation process of this problem.}
\end{tabular} \\ \hline

\multirow{6}{*}{\textbf{Reasoning Path}} 
& \textbf{PS} \\
& Plan:  1. Find the combined sq ft of the bedroom and bathroom.  2. Add 200 sq ft to the combined sq ft from step 1.  3. Multiply the combined sq ft from step 2 by 2.   Solution:  1. 309 sq ft (bedroom) + 150 sq ft (bathroom) = 459 sq ft  2. 459 sq ft + 200 sq ft = 659 sq ft  3. 659 sq ft x 2 = 1318 sq ft   Therefore, the new room will have \sethlcolor{yellow}\hl{1318} sq ft. \textcolor{red}{\XSolidBrush} \\ \cline{2-2} 
& \textbf{PS + I$^3$C} \\ 
& First, we need to calculate the combined area of Holden's bedroom and bathroom, which is equal to 309 sq ft (bedroom) + 150 sq ft (bathroom) = 459 sq ft.   Next, we need to double this combined area to get the total area of the new room. This can be done by multiplying 459 sq ft (the total sq ft of Holden's bedroom and bathroom combined) by 2, which gives us 918 sq ft.   Therefore, the new room will have \sethlcolor{yellow}\hl{918} sq ft. The basketball field of Tom is 200 sq ft larger than Holden's bathroom so it is not relevant to the calculation process of this problem. \textcolor{red}{\Checkmark} \\ \cline{2-2} 
& \textbf{I$^3$C-Select} \\ 
& Holden's bedroom is 309 sq ft and his bathroom is 150 sq ft. The total sq ft of Holden's bedroom and bathroom combined is 459 sq ft. We need to find a space that is twice as large as Holden's bedroom and bathroom combined. Therefore, the new room will be 2 x 459 sq ft = \sethlcolor{yellow}\hl{918} sq ft. The condition The basketball field of Tom is 200 sq ft larger than Holden's bathroom is not relevant to the calculation process of this problem, as it does not involve any calculation involving Holden's bedroom and bathroom. \textcolor{red}{\Checkmark}  \\

\hline                                
\end{tabularx} }
\label{tab:case_3}
\end{table*}

\begin{table*}[t]
\centering
\caption{Example outputs by Instruct-CoT, Instruct-CoT + I$^3$C and I$^3$C-Select. Irrelevant conditions in the problem description is highlighted with red color, and the LLM-generated final answer is highlighted with yellow color.}
\renewcommand\arraystretch{1.35}
\resizebox{\textwidth}{!}{
\begin{tabularx}{\textwidth}{p{2.5cm}|X}
\hline   

\textbf{Problem} & James buys 6 t-shirts for 50\% off. They each cost \$20. \sethlcolor{pink}\hl{Emma works at the mall with an hourly salary of \$10.} How much did James pay? \\ \hline

\textbf{Gold Answer} & $(20 / 2) \times 6 = 60$ \\ \hline

\textbf{Verification Outputs} & \begin{tabular}[c]{@{}p{12.8cm}@{}}
$\blacksquare$ First, James paid 50\% off the normal price for the 6 t-shirts, which would normally cost \$20 each. This means he paid \$10 per t-shirt.  So, James paid \$60 total for the 6 t-shirts.  \textcolor{teal!70}{The condition James buys 6 t-shirts for 50\% off is relevant to the calculation process, as it determined the amount that James paid for the 6 t-shirts.  Emma's hourly salary of \$10 is not relevant to the calculation process of this problem.}  \\
$\blacksquare$ James bought 6 t-shirts, so he paid for 6 times \$20. Then, he got 50\% off, so 50\% of \$20 is \$10, so he paid for 6 times \$10. \textcolor{teal!70}{The condition They each cost \$20 is relevant to the problem, because it tells us how much James paid for each t-shirt. So, the answer is James paid \$60.}  \\
$\blacksquare$ James bought 6 t-shirts at 50\% off, which means the t-shirts were 50\% cheaper than their original price. This means that each t-shirt cost \$20. Multiplying 6 t-shirts by \$20, James paid a total of \$120 for the 6 t-shirts.   \textcolor{teal!70}{The condition that Emma works at the mall with an hourly salary of \$10 is not relevant to the calculation process of this problem.}
\end{tabular} \\ \hline

\multirow{6}{*}{\textbf{Reasoning Path}} 
& \textbf{Instruct-CoT} \\
& Since James bought 6 t-shirts for 50\% off, the price of each t-shirt is reduced to \$20. Therefore, James paid \sethlcolor{yellow}\hl{\$120} in total for the 6 t-shirts. Emma’s hourly salary of \$10 is irrelevant to the question. \textcolor{red}{\XSolidBrush} \\ \cline{2-2} 
& \textbf{Instruct-CoT + I$^3$C} \\ 
& James bought 6 t-shirts that were 50\% off, so the original price of each t-shirt was \$20. Therefore, James paid \$10 for each t-shirt, and he bought 6 t-shirts, so he paid \sethlcolor{yellow}\hl{\$60} in total. The condition that Emma works at the mall with an hourly salary of \$10 is not relevant to the calculation process of this problem. \textcolor{red}{\Checkmark} \\ \cline{2-2} 
& \textbf{I$^3$C-Select} \\ 
& First, James paid 50\% off the normal price for the 6 t-shirts, which would normally cost \$20 each. This means he paid \$10 per t-shirt.  So, James paid \sethlcolor{yellow}\hl{\$60} total for the 6 t-shirts.  The condition James buys 6 t-shirts for 50\% off is relevant to the calculation process, as it determined the amount that James paid for the 6 t-shirts.  Emma's hourly salary of \$10 is not relevant to the calculation process of this problem. \textcolor{red}{\Checkmark}  \\

\hline                                
\end{tabularx} }
\label{tab:case_4}
\end{table*}

\begin{table*}[t]
\centering
\caption{Example outputs by Manual-CoT, Manual-CoT + I$^3$C and I$^3$C-Select. Irrelevant conditions in the problem description is highlighted with red color, and the LLM-generated final answer is highlighted with yellow color.}
\renewcommand\arraystretch{1.35}
\resizebox{\textwidth}{!}{
\begin{tabularx}{\textwidth}{p{2.5cm}|X}
\hline   

\textbf{Problem} & James collects all the fruits from his 2 trees. Each tree has 20 plants. Each plant has 1 seed and he plants 60\% of those. \sethlcolor{pink}\hl{James' sister is 10 years old.} How many trees did James plant? \\ \hline

\textbf{Gold Answer} & $2 \times 20 \times 60\% = 24$ \\ \hline

\textbf{Verification Outputs} & \begin{tabular}[c]{@{}p{12.8cm}@{}}
$\blacksquare$ James collected all the fruits from his 2 trees. This means he did not plant any new trees.  Each tree has 20 plants. This means James had 40 plants in total.  Each plant has 1 seed. This means James had 40 seeds in total.  He planted 60\% of those seeds. This means he planted 24 seeds.   Therefore, James planted 24 seeds, but he did not plant any new trees. \textcolor{teal!70}{The condition James collects all the fruits from his 2 trees is relevant to the calculation process because it indicates that James did not plant any new trees, only seeds.}  \\
$\blacksquare$ First, James collected all the fruits from his two trees. Each tree has 20 plants, so James collected 40 plants in total.  Second, each plant has 1 seed, so James collected 40 seeds in total.  Third, he planted 60\% of those seeds, so James planted 24 seeds in total.  Finally, we can conclude that James planted two trees, since each tree had 20 plants and he planted 24 seeds in total.  So, to answer the question, James planted two trees.  \textcolor{teal!70}{Yes, the condition Each tree has 20 plants is relevant to the calculation process of the problem How many trees did James plant?, since it helps us determine how many plants he collected in total and how many seeds he planted in total.} \\
$\blacksquare$ James collects all the fruits from his 2 trees. Each tree has 20 plants. Each plant has 1 seed.  This means there are 40 seeds in total.   James then plants 60\% of those, which means he plants 24 seeds.   \textcolor{teal!70}{The condition James' sister is 10 years old is not relevant to the calculation process of this problem.   Therefore, James planted 24 trees.}
\end{tabular} \\ \hline

\multirow{6}{*}{\textbf{Reasoning Path}} 
& \textbf{Manual-CoT} \\
& James collected all the fruits from his 2 trees. Each tree has 20 plants. Each plant has 1 seed. He planted 60\% of those. 60\% is 0.6, so he planted 0.6 x 20 = 12 of those. James' sister is 10 years old. This has nothing to do with the question, so the answer is still \sethlcolor{yellow}\hl{12}. \textcolor{red}{\XSolidBrush} \\ \cline{2-2} 
& \textbf{Manual-CoT + I$^3$C} \\ 
& James collected all the fruits from his two trees, so he has 40 plants. Each plant has 1 seed, so he has 40 seeds. He plants 60\% of those, so he plants 24 seeds. Therefore, James planted \sethlcolor{yellow}\hl{24} trees. The condition James' sister is 10 years old is not relevant to the calculation process of this problem. \textcolor{red}{\Checkmark} \\ \cline{2-2} 
& \textbf{I$^3$C-Select} \\ 
& First, James collected all the fruits from his two trees, so he has 40 plants.   Second, each plant has 1 seed, so he has 40 seeds.   Third, he plants 60\% of those, so he plants 24 seeds.   Therefore, James planted 24 trees.   The condition that James' sister is 10 years old is not relevant to the calculation process of this problem. The relevant information to answer this question is that James collected all the fruits from his two trees, each tree has 20 plants, and he plants 60\% of those. To calculate how many trees James planted, we need to calculate the number of seeds he planted. So the answer would be: \sethlcolor{yellow}\hl{24} trees. \textcolor{red}{\Checkmark}  \\

\hline                                
\end{tabularx} }
\label{tab:case_5}
\end{table*}

\begin{table*}[t]
\centering
\caption{Example outputs by Auto-CoT, Auto-CoT + I$^3$C and I$^3$C-Select. Irrelevant conditions in the problem description is highlighted with red color, and the LLM-generated final answer is highlighted with yellow color.}
\renewcommand\arraystretch{1.35}
\resizebox{\textwidth}{!}{
\begin{tabularx}{\textwidth}{p{2.5cm}|X}
\hline   

\textbf{Problem} & Hamza has several empty buckets of different sizes, holding either 3, 5, or 6 liters. She fills the 5-liter bucket and pours as much as she can into the 3-liter bucket. Then, she pours the remainder into the 6-liter bucket. \sethlcolor{pink}\hl{Jack drinks 7 liters of milk every day.} How much more water, in liters, can Hamza put into the 6-liter bucket, without overflowing? \\ \hline

\textbf{Gold Answer} & $6 - (5 - 3) = 4$ \\ \hline

\textbf{Verification Outputs} & \begin{tabular}[c]{@{}p{12.8cm}@{}}
$\blacksquare$ First, Hamza fills the 5-liter bucket and then pours as much as she can into the 3-liter bucket. Since the 3-liter bucket has a capacity of 3 liters, she can pour all 5 liters into it.  Then, she pours the remainder (2 liters) into the 6-liter bucket. That means she has put a total of 7 liters into the 6-liter bucket.  Now, Jack drinks 7 liters of milk every day. That means the 6-liter bucket is emptied every day.  Therefore, Hamza can put up to 6 liters more water, in liters, into the 6-liter bucket, without overflowing.   \textcolor{teal!70}{Yes, the condition Hamza has several empty buckets of different sizes, holding either 3, 5, or 6 liters. is relevant to the calculation process of the problem. This condition gives us the capacity of each bucket, which is necessary to calculate the amount of water Hamza can put into the 6-liter bucket without overflowing.}  \\
$\blacksquare$ We can calculate that Hamza can put an additional 3 liters of water into the 6-liter bucket, without overflowing. This is because the 6-liter bucket can hold a maximum of 6 liters of water, and it currently holds 3 liters of water.  \textcolor{teal!70}{The condition Jack drinks 7 liters of milk every day is not relevant to the calculation process of this problem.} \\
\end{tabular} \\ \hline

\multirow{6}{*}{\textbf{Reasoning Path}} 
& \textbf{Auto-CoT} \\
& The 5-liter bucket is filled. She pours as much as she can into the 3-liter bucket. That leaves 2 liters in the 5-liter bucket. She then pours the remainder into the 6-liter bucket. That leaves 2 liters in the 6-liter bucket. Jack drinks 7 liters of milk every day, so she has to leave 7 liters of water for him. Therefore, she can put 5 liters of water in the 6-liter bucket without overflowing. The answer is \sethlcolor{yellow}\hl{5}. \textcolor{red}{\XSolidBrush} \\ \cline{2-2} 
& \textbf{Auto-CoT + I$^3$C} \\ 
& Hamza first fills the 5-liter bucket. She then pours as much as she can into the 3-liter bucket, filling it to capacity. The remaining 2 liters of water from the 5-liter bucket are then poured into the 6-liter bucket, leaving 4 liters of space for more water (6 liters - 2 liters = 4 liters). Hamza can add 4 more liters of water into the 6-liter bucket without overflowing. The answer is \sethlcolor{yellow}\hl{4}.  \textcolor{red}{\Checkmark}   \\ \cline{2-2} 
& \textbf{I$^3$C-Select} \\ 
& Hamza has a 5-liter bucket. This is relevant to the calculation process as it tells us the size of the bucket Hamza has. She fills the 5-liter bucket and pours as much as she can into the 3-liter bucket. This is relevant to the calculation process as it tells us how much water Hamza is transferring from the 5-liter bucket to the 3-liter bucket. The 3-liter bucket can only hold 3 liters of water, so Hamza would be able to pour 3 liters of water from the 5-liter bucket into the 3-liter bucket. She then pours the remainder of the water (2 liters) into the 6-liter bucket. This is relevant to the calculation process as it tells us how much water Hamza is transferring from the 5-liter bucket to the 6-liter bucket. The 6-liter bucket currently has 2 liters of water in it, so it can hold 4 more liters of water without overflowing. So the answer would be \sethlcolor{yellow}\hl{4} liters. \textcolor{red}{\Checkmark}  \\

\hline                                
\end{tabularx} }
\label{tab:case_6}
\end{table*}

\end{document}